\documentclass[10pt,twocolumn,letterpaper]{article}

\usepackage[pagenumbers]{cvpr} 

\usepackage{graphicx}
\usepackage{amsmath}
\usepackage{amssymb}
\usepackage{booktabs}
\usepackage{booktabs}
\usepackage{algorithm}
\usepackage{soul}
\usepackage{algpseudocode}
\usepackage{multirow}
\usepackage{xspace}
\usepackage{tabularx}
\usepackage{makecell}
\usepackage{xspace}
\usepackage{siunitx}
\usepackage[pagebackref,breaklinks,colorlinks]{hyperref}

\newcommand{\PreserveBackslash}[1]{\let\temp=\\#1\let\\=\temp}
\newcolumntype{C}[1]{>{\PreserveBackslash\centering}p{#1}}
\newcolumntype{R}[1]{>{\PreserveBackslash\raggedleft}p{#1}}
\newcolumntype{L}[1]{>{\PreserveBackslash\raggedright}p{#1}}

\def\cvprPaperID{610} 
\def\confName{CVPR}
\def\confYear{2023}

\newcommand{\ours}{NIRVANA\xspace}
\newcommand{\Sec}{Section\xspace}
\newcommand{\Fig}{Fig.\xspace}

\makeatletter
\DeclareRobustCommand\onedot{\futurelet\@let@token\@onedot}
\def\@onedot{\ifx\@let@token.\else.\null\fi\xspace}

\newcommand{\bw}{\boldsymbol{W}}

\newcommand{\bwt}{\widetilde{\bw}}

\makeatother

\begin{document}

\title{\ours: Neural Implicit Representations of Videos with \\[0.1em]
Adaptive Networks and Autoregressive Patch-wise Modeling}
\author{\normalsize
Shishira R Maiya$^{1,2}$\thanks{First two authors contributed equally\newline}, Sharath Girish$^1$\footnotemark[1], Max Ehrlich$^1$, Hanyu Wang$^1$, Kwot Sin Lee$^2$,\\
\normalsize  Patrick Poirson$^2$, Pengxiang Wu$^2$,
\normalsize Chen Wang$^2$, Abhinav Shrivastava$^1$ \\
\normalsize
$^1$University of Maryland \qquad $^2$Snap Inc\\
{\tt \footnotesize \{maxehr,sgirish,shishira,hwang\}@umd.edu}\\ {\tt \footnotesize \{klee6, ppoirson, pwu,cwang6\}@snapchat.com} {\tt \footnotesize abhinav@cs.umd.edu}
}

\maketitle

\begin{abstract}

  Implicit Neural Representations (INR) have recently shown to be powerful tool for high-quality video compression. However, existing works are limiting as they do not explicitly exploit the temporal redundancy in videos, leading to a long encoding time. Additionally, these methods have fixed architectures which do not scale to longer videos or higher resolutions. To address these issues, we propose \ours, which treats videos as groups of frames and fits separate networks to each group performing patch-wise prediction. 
   The video representation is modeled autoregressively, with networks fit on a current group initialized using weights from the previous group's model. To further enhance efficiency, we perform quantization of the network parameters during training, requiring no post-hoc pruning or quantization. When compared with previous works on the benchmark UVG dataset, \ours improves encoding quality from 37.36 to 37.70 (in terms of PSNR) and the encoding speed by 12$\times$, while maintaining the same compression rate. In contrast to prior video INR works which struggle with larger resolution and longer videos, we show that our algorithm is highly flexible and scales naturally due to its patch-wise and autoregressive designs. Moreover, our method achieves variable bitrate compression by adapting to videos with varying inter-frame motion. \ours achieves 6$\times$ decoding speed and scales well with more GPUs, making it practical for various deployment scenarios.
\end{abstract}
\vspace{-0.5em}

\begin{figure}
    \centering
    \includegraphics[width=0.9\linewidth]{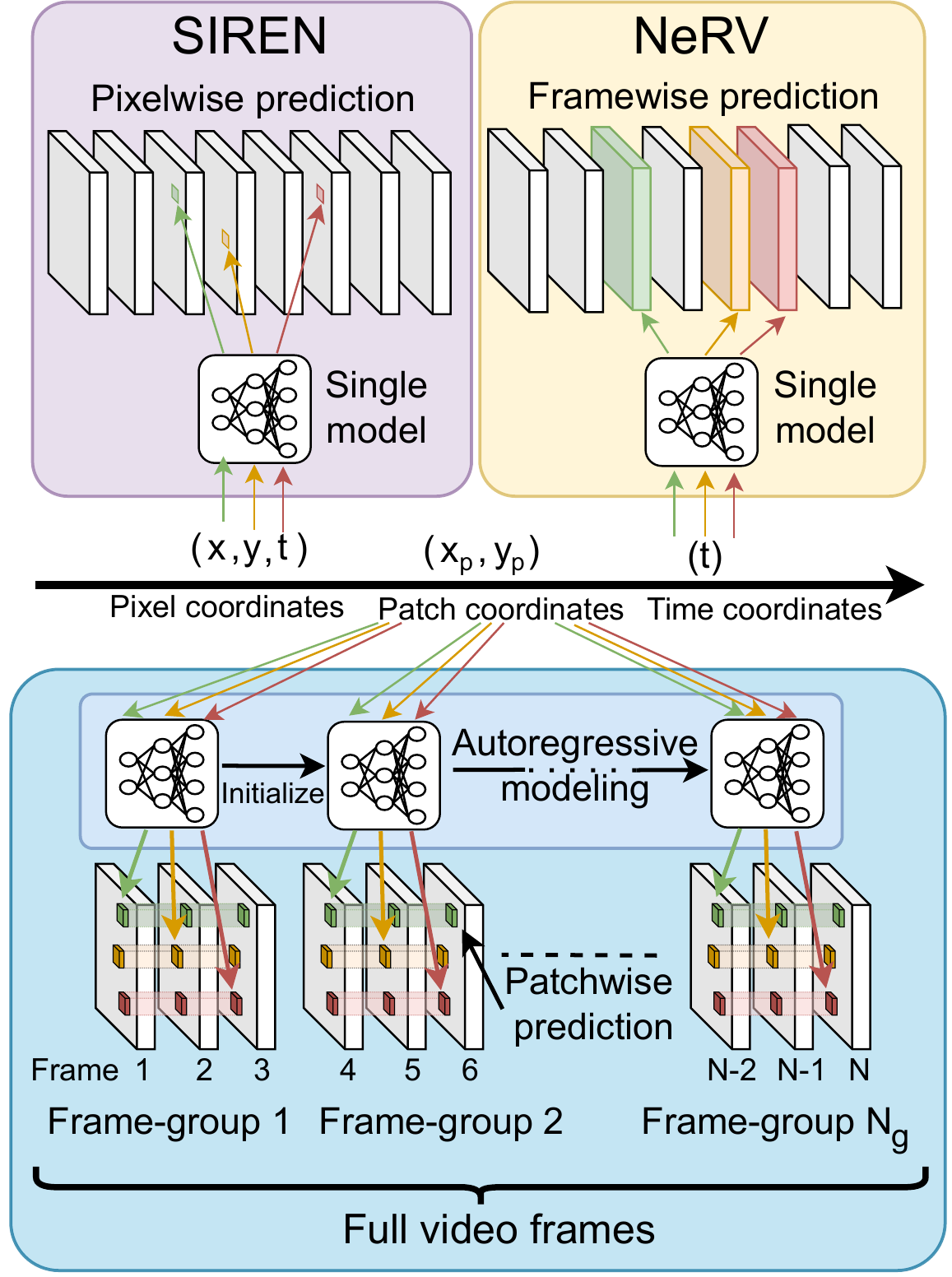}
    \caption{\textbf{Overview of \ours:} Prior video INR works perform either pixel-wise or frame-wise prediction. We instead perform spatio-temporal patch-wise prediction and fit individual neural networks to groups of frames (clips) which are initialized using networks trained on the previous group. Such an autoregressive patch-wise approach exploits both spatial and temporal redundancies present in videos while promoting scalability and adaptability to varying video content, resolution or duration.}
    \vspace{-1.3em}
\end{figure}

\section{Introduction}
\label{sec:intro}

In the information age today, where petabytes of content is generated and consumed every hour, the ability to compress data fast and reliably is important. Not only does compression make data cheaper for server hosting, it makes content accessible to population/regions with low-bandwidth. Conventionally, such compression is achieved through codecs like JPEG~\cite{125072} for images and HEVC~\cite{sze2014high}, AV1~\cite{chen2018overview} for videos, each of which compresses data via targetted hand-crafted algorithms. These techniques achieve acceptable trade-offs, leading to their widespread usage. 

With the rise of deep learning, machine learning-based codecs~\cite{balle2016end,balle2018variational,minnen2018joint,theis2017lossy} showed that it is possible to achieve better performance in some aspects than conventional codecs. However, these techniques often require large networks as they attempt to generalize to compress all data from the distribution. Furthermore, such generalization is contingent on the training dataset used by these models, leading to poor performance for Out-of-Distribution (OOD) data across different domains~\cite{zhang2021out} or even when the resolution changes~\cite{cao2022oodhdr}. This greatly limits its \textit{real-world practicality} especially if the input data to be compressed is significantly different from what the model has seen during training.
In recent years, a new paradigm, Implicit Neural Representations (INR), emerged to solve the drawbacks of model-learned compression methods. Rather than attempting to generalize to all data from a particular distribution, its key idea is to train a network that specifically fits to a signal, which can be an image~\cite{sitzmann2020implicit}, video~\cite{chen2021nerv}, or even a 3D scene~\cite{mildenhall2020nerf}. With this conceptual shift, a neural network is no longer just a predictive tool, rather it is now an efficient storage of data. 
Treating the \textit{neural network as the data}, INR translate the data compression task to that of model compression. Such a continuous function mapping further benefits downstream tasks such as image super-resolution~\cite{li2022efficient}, denoising~\cite{muller2022instant}, and inpainting~\cite{sitzmann2020implicit}.

Despite these advances, videos vary widely in both spatial resolutions and temporal lengths, making it challenging for networks to encode videos in a practical setting. Towards solving this task, an early method, SIREN~\cite{sitzmann2020implicit}, attempted to learn a direct mapping from 3D spatio-temporal coordinates of a video to each pixel's color values. While simple, this is computationally inefficient and does not factor in the spatio-temporal redundancies within the video. Later, NeRV~\cite{chen2021nerv} proposed to instead map 1D temporal coordinates in the form of frame indices directly to generate a whole frame. While this improves the reconstruction quality, such a mapping still does not capture the temporal redundancies between frames as it treats each frame individually as a separate image encoding task. 
Finally, mapping only the temporal coordinate also means one would need to modify the architecture in order to encode videos of different spatial resolutions.

To address the above issues, we propose \ours, a method that exploits spatio-temporal redundancies to encode videos of arbitrary lengths and resolutions. Rather than performing a pixel-wise prediction (\textit{e.g.}, SIREN) or a whole-frame prediction (\textit{e.g.}, NeRV), we predict \textit{patches}, which allows our model to adapt to videos of different spatial resolutions without modifying the architecture. Our method takes in the centroids of patches (patch coordinates) $(x_p,y_p)$ as inputs and outputs a corresponding patch volume. Since patches can be arranged for different resolutions, we do not require any architectural modification when the input video resolution changes. Furthermore, to exploit the temporal nature of videos, we propose to train individual, small models for each group of video frames (``clips") in an autoregressive manner: the model weights for predicting each frame group is initialized from the weights of the model for the previous frame group. Apart from the obvious advantage that we can scale to longer sequences without changing the model architecture, this design exploits the temporal nature of videos that, intuitively, frame groups with similar visual information (\textit{e.g.}, static video frames) would have similar weights, allowing us to further perform residual encoding for greater compression gains. This \textit{adaptive} nature, that static videos gain greater compression than dynamic ones, is a big advantage over NeRV where the compression for identical frames remain fixed as it models each frame individually. To obtain further compression gains, we employ recent advances in the literature to add entropy regularization for quantization, and encode model weights for each video during training \cite{Girish2022LilNetXLN}. This further adapts the compression level to the complexity of each video, and avoids any post-hoc pruning and fine-tuning as in NeRV, which can be slow.

Finally, we show that despite its autoregressive nature, our model is linearly parallelizable with the number of GPUs by chunking each video into disjoint groups to be processed. This strategy largely improves the speed while maintaining the superior compression gains of our method, making it practical for various deployment scenarios.

We evaluate \ours on the benchmark UVG dataset \cite{10.1145/3339825.3394937}. We show that \ours reaches the same levels of PSNR and Bits-per-Pixel (BPP) compression rate with almost $12\times$ the encoding speed of NeRV. We verify that our algorithm adapts to varying spatial and temporal scales by providing results on videos in the UVG dataset with 4K resolution at 120fps, as well as on significantly longer videos from the YouTube-8M dataset, both of which are challenging extensions which have not been attempted on for this task. We show that our algorithm outperforms the baseline with much smaller encoding times and that it naturally scales with no performance degradation. We conduct ablation studies to show the effectiveness of various components of our algorithm in achieving high levels of reconstruction quality and understand the sources of improvements.

Our contributions are summarized below:
\begin{itemize}
\setlength\itemsep{-.3em}
    \item We present \ours, a patch-wise autoregressive video INR framework which exploits both spatial and temporal redundancies in videos to achieve high levels of encoding speedups (12$\times$) at similar reconstruction quality and compression rates.
        \item We achieve a 6$\times$ speedup in decoding times and scale well with increasing number of GPUs, making it practical in various deployment scenarios.
    \item Our framework adapts to varying video resolutions and lengths without performance degradations. Different from prior works, it achieves adaptive bitrate compression based on the amount of inter-frame motion for each video.
\end{itemize}

\section{Related Work}
\noindent\textbf{Implicit Neural Representations (INR)} are a novel family of methods designed to map a set of coordinates to a specific signal - such as a single image or video - using a neural network as a function for such mappings. 
SIREN \cite{sitzmann2020implicit} demonstrates that by utilizing periodic activation functions in MLPs, such a function can be fit and used to map a wide array of signals, including images, 3D shapes and videos.
As an alternative, \cite{tancik2020fourfeat} shows than an INR network with standard activations can be trained by utilizing random Fourier features. \cite{martel2021acorn} and \cite{saragadam2022miner} propose adaptive block-based approaches whose complexity mirrors the underlying signals. Frequency-based approaches are proposed in~\cite{Fathony2021MultiplicativeFN,lindell2022bacon,shekarforoush2022residual} that enable multi-scale representations. 
The first image specific INR method is COIN~\cite{dupont2021coin}, which is extended to encode multiple images through network modulations in COIN++~\cite{dupont2022coin++}. A method for learning local implicit functions is proposed in~\cite{chen2020learning} that results in smoother super-resolution outputs. 
Several methods~\cite{tancik2020meta,strumpler2021implicit,chen2022transinr} have explored meta-learning approaches to reduce the long encoding times of image INRs;~\cite{lee2021meta} further shows that directly initializing a meta-sparse network not only gives a good initialization but also helps with model compression.

\noindent\textbf{INR for videos.} Despite the significant advances in INR methods for image compression, videos present a more challenging task for INR methods. 
For example, if we naively add time as an extra dimension to image-based methods, such as in SIREN~\cite{sitzmann2020implicit}, the resulting outputs are grainy. In~\cite{rho2022neural,zhang2021implicit}, INR methods that utilize flow-based information to encode videos are introduced; however, they cannot scale beyond short low-resolution videos. NeRV~\cite{chen2021nerv} is the first method to scale video compression using INRs. They modify the implicit mapping function to learn a direct mapping from a video frame index to an entire frame. Further extensions of NeRV, such as patch-based versions~\cite{bai2022ps,li2022nerv}, provide minor improvements over the original architecture. Despite good reconstruction, the problems of long encoding times, the lack of inter-frame encoding, and the inability to adapt to video content act as major drawbacks for widespread adoptions.

\noindent\textbf{Model compression} is typically achieved by pruning or quantization of network weights. A plethora of works perform model pruning with minimal loss of performance \cite{frankle2018lottery,frankle2019stabilizing,girish2021lottery,gale2019state}. Pruned models contain a majority of zeros and can be stored in sparse matrix formats \cite{ivanovsten} for reduction in model size. Alternatively, quantization works \cite{stock2019and,fan2020training,tung2018clip,lin2017towards} to reduce the number of bits needed to store each model weight, resulting in reduced disk space.
As implicit neural networks represent the data using their model weights, data compression translates to model compression. 
In this work, we adopt the works of \cite{oktay2019scalable,Girish2022LilNetXLN} for model compression by maintaining a set of quantized parameters during training which are then stored on disk. These recent methods have shown to achieve high levels of compression through entropy regularization without sacrificing downstream network performance. 
We perform quantization and model compression \textit{during training}, unlike the 
post-hoc pruning and quantization in NeRV \cite{chen2021nerv}.

\section{Approach}
\label{sec:approach}

\subsection{Background}
\label{ssec:background}
Given a video $\boldsymbol{V} \in \mathbb{R}^{N\times H\times W\times 3}$ consisting of $N$ frames, each with spatial resolution $H\times W$, an INR defines a mapping from the spatio-temporal coordinate  $X = (x, y, t); X \in \mathbb{R}^3$ to the pixel value $p \in \mathbb{R}^3$. Thus, it implicitly represents a parameterized function $h_\theta:\mathbb{R}^3 \rightarrow \mathbb{R}^3$ parameterized by $\theta$. The function is typically trained by minimizing the MSE-loss $\vert\vert h_{\theta}(X) - \boldsymbol{V}\vert\vert_{2}$. 
While this is a straightforward extension of image-based INR methods to the spatio-temporal domain, it fails to exploit the spatial and temporal consistency in videos. Pixel-wise prediction leads to redundant computation and long encoding times while also producing blurry outputs\cite{sitzmann2020implicit},\cite{dupont2021coin}. To mitigate this issue, NeRV \cite{chen2021nerv} proposes to directly encode the frame index $t \in \mathbb{R}$ as a positional embedding input to a model which outputs the entire image frame $\mathbb{R}^{H\times W\times 3}$. NeRV consists of several MLP layers followed by convolution layers which upsamples the latent representation to the target frame's spatial resolution. While this formulation improves upon the naive pixel-based formulation, it does not adapt to arbitrary resolutions, and does not capture the temporal dependencies between frames as it effectively acts as only an image encoder for each frame.
\begin{figure*}[!ht]
    \centering
    \includegraphics[width=\textwidth]{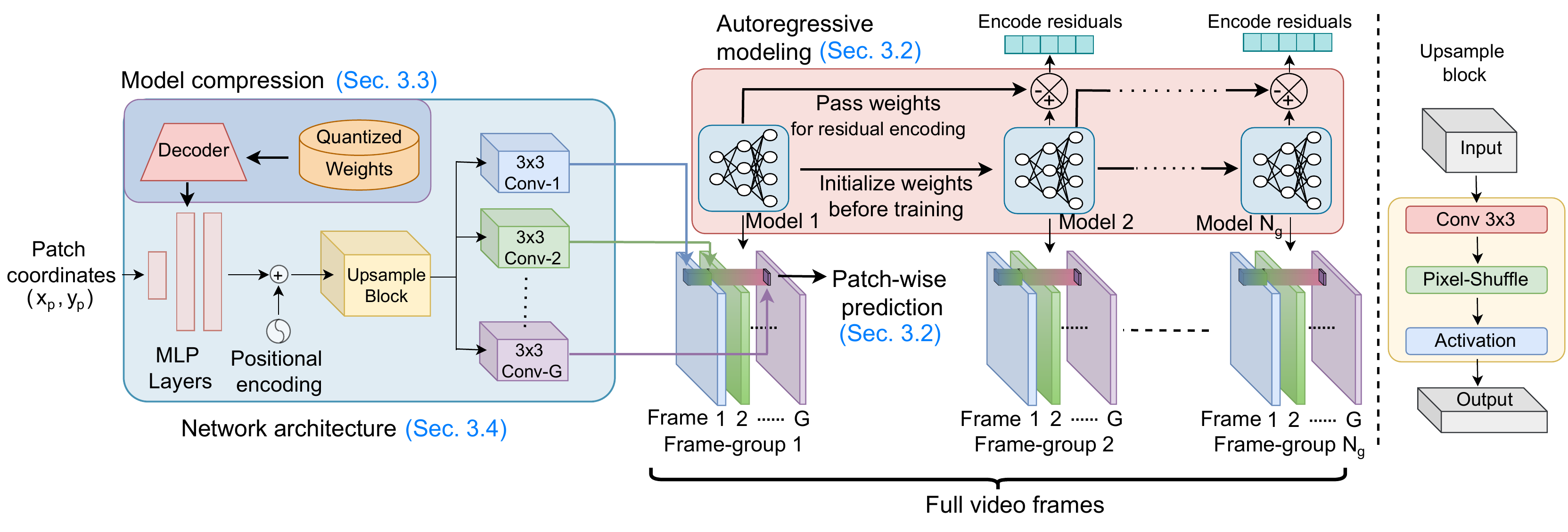}
    \vspace{-0.3in}
    \caption{\textbf{Overview of \ours:} We propose an autoregressive video INR framework which performs patch-wise prediction of groups of frames by fitting separate networks to each group. Each network is initialized with the previous group's network weights. Our architecture consists of several SIREN MLP layers followed by an upsampling block (right). It takes patch coordinates as inputs and outputs patches across a group of G frames. We maintain a set of quantized weights which are decoded to obtain the network weights. Post training, we encode weight residuals which are the difference between the quantized weights of the current and previous group's network.}
    \label{fig:arch}
\end{figure*}

\subsection{Autoregressive Patch-wise Modeling}  
\label{ssec:autoreg}
\paragraph{Patch prediction.} The two dominant INR paradigms for video encodings, SIREN \cite{sitzmann2020implicit} and NeRV \cite{chen2021nerv} represents two extreme ends of a spectrum: the former predicts every pixel in a video volume independently, while the latter predicts the pixels for a single frame simultaneously. Through pixel-wise prediction, SIREN accommodates varying the output image's spatial resolution but does not exploit the spatial consistency of the image. In contrast, while NeRV exploits the spatial consistency of an image through convolutions, it cannot vary the image's spatial resolution. We adopt a middle ground between these two extremes by adopting a modeling approach that instead predicts \textit{patches} of an image. This gives us the best of both approaches, since our model utilizes the spatial consistency of an image while still naturally scaling to varying image resolutions. 

We push further in this direction by exploiting the temporal consistency in videos to predict a \textit{volume of patches} across neighboring frames. We thus predict a patch group $\boldsymbol{P}\in \mathbb{R}^{G\times H_p \times W_p \times 3}$, where G is the number of frames in a group and $(H_p,W_p)$ is the patch size. This method enables us to reduce the amount of redundant computation in both the spatial and temporal dimensions, leading to significantly shorter encoding times compared to NeRV or SIREN.
\vspace{-1.5mm}
\paragraph{Autoregressive networks.} While it is straightforward to input a 3D patch coordinate $(x_p,y_p,g)$ (where $(x_p,y_p)$ are the patch centroids within a frame and $g$ is the frame index within each group) and output a patch volume using a single network, it still suffers from adaptability to varying video content, resolutions, or durations as mentioned in \Sec\ref{ssec:background}.
To overcome these challenges, we propose to autoregressively fit separate networks to each frame group. Each network is fed with the same set of inputs, namely the centroids of patches $(x_p,y_p)$, and it outputs the corresponding 3D patch volume of the group. Every subsequent network is finetuned from the previous one, leading to shorter encoding times. As video content does not change much over a short time period of multiple frames, the difference in weights (or weight residuals) after fine-tuning is small. Thus, encoding the weight residuals instead of the weights themselves leads to higher compression rates.
The design of the algorithm \ref{alg:vid_INR} allows us to encode multiple chunks of a long video in a parallel manner, a key feature missing from existing methods. This means \ours can scale linearly with the number of GPUs without any drop in performance (see \Sec\ref{ssec:gpu_parallel}). 
\subsection{Model Compression and Weight Storage}
\label{ssec:model_comp}
Since the network weights are the latent representations for the video, network size directly translates to bitrate of the video encoding. To reduce network size, we adapt existing works which perform model weights/latent representation compression \cite{oktay2019scalable,Girish2022LilNetXLN}. For a weight parameter $\boldsymbol{W} \in \theta$, where $\theta$ is the set of model parameters, we maintain a corresponding quantized latent weight $\boldsymbol{\widetilde{W}}$. The continuous weight parameter $\boldsymbol{W}$ is then obtained as $\boldsymbol{W} = f_\phi(\boldsymbol{\widetilde{W}})$ where $f_\phi$ is a 
learned linear transform. The entire setup is trained end to end, without any post-hoc quantization and fine-tuning.
As previously explained, we encode the quantized latent residuals instead of the latents themselves to achieve higher levels of compression.
We encode these residuals using arithmetic coding \cite{witten1987arithmetic}, a lossless entropy-based compression algorithm. In order to encourage the latents to have lower entropy, we add an entropy regularization term to our loss function. This term encourages the network to have a lower entropy and hence a lower bitrate.
When decoding, each weight is obtained sequentially by cumulatively adding residuals. This approach helps in making the bitrate of \ours adaptive to the video content: for frame-groups that have little motion, they are already closer to convergence and thus have small differences in their network weights, leading to sparser residuals and subsequently, lower bitrates. This feature is missing in other methods as the models are fixed for a given video. 
\subsection{Network Architecture}
\label{ssec:arch}
In this section, we describe the network architecture which is used for each frame group, as illustrated in \Fig\ref{fig:arch}. For a group of $G$ frames, we segment patch volumes of shape $(H_p,W_p,G)$. The input to the network is the 2D patch coordinate $(x_p,y_p) \in \mathbb{R}^2$ and the output is the corresponding RGB patch volume $\boldsymbol{P} \in \mathbb{R}^{G\times H_p \times W_p \times 3}$. We stack multiple MLP layers with SIREN activations to obtain an output feature representation vector $s_p \in \mathbb{R}^d$ of dimension $d$. We replicate $s_p$ by $G$ times, and add positional encoding vectors based on the position of the frame within the group, using the following embedding function:
\begin{align}
    \tau(t,2i) = \sin\left(\frac{t}{f^{2i}}\right) \tau(t,2i+1) = \cos\left(\frac{t}{f^{2i}}\right), i \in [0,d)
\end{align}
where $t \in [0,G)$ represents the position of the frame within the group of $G$ frames. We then add a decoder block as in \cite{chen2021nerv} followed by a $3\times3$ convolutional layer to output $G$ patches. 
For a video with $N$ frames, we segment it into $N_g$ frame-groups with each group consisting of $G$ frames ($N=N_g\times G$). For the $g^\text{th}$ frame-group ($g \in [0,N_g)$), the corresponding network is represented as $h_{\theta_g}$ consisting of parameters $\theta_g$. The overall loss objective for training the network for the $g^\text{th}$ frame-group is therefore
\begin{align}
    \mathcal{L}_g = \mathcal{L}_\text{mse}(h_{\theta_g}(p),v_g) +\lambda_I \mathcal{L}_\text{ent}(\theta_g)
    \label{eqn:mse}
\end{align}
where $p$ represents patch grid coordinates and $v_g$ means the corresponding ground-truth frame-group pixel values. $\mathcal{L}_\text{ent}(\theta_g)$ represents the entropy regularization loss on the model parameters $\theta_g$. It is weighed by the coefficient $\lambda_I$ controlling the rate-distortion trade-off for reconstructing the frame groups.

\begin{algorithm}[t]
\caption{Sequential Video INR}\label{alg:vid_INR}
\begin{algorithmic}[1]
\State \textbf{Init} Randomly initialize network $h_{\theta_0}$ with initial weights $\theta_0$ and training iterations $T$. The video contains $N$ frames segmented into $N_{g}$ frame groups of size G each.
\For{$g$ in $0,1,2,\cdots, N_{g} - 1$ } 
\If{$g=0$}
    \State Train $h_{\theta_0}$ for $T$ iterations for all patches
    \State Store weights $h_{\theta_0}$
\Else
    \State Initialize weights $h_{\theta_k} \leftarrow h_{\theta_{k-1}}$
    \State Finetune $h_{\theta_k}$ for $T_{r}$ iterations for all patches
    \State Store quantized latent of residuals $h_{\theta_k}-h_{\theta_{k-1}}$
\EndIf
\EndFor
\end{algorithmic}
\end{algorithm}

\section{Experiments}
\subsection{Datasets and Implementation Details}

The standard benchmark UVG dataset \cite{10.1145/3339825.3394937} is used to compare our approach \ours with prior video INR works. Following similar setups~\cite{chen2021nerv}, our approach is evaluated on 7 videos from the dataset at 1080p resolution (UVG-HD) and 120 fps with 6 videos consisting of 600 frames and 1 with 300 frames. To show our scalability for higher resolution videos, we show results for the 7 videos at 4K resolution (UVG-4K) as well. We additionally include a video from the Youtube-8M (see Appendix) \cite{abu2016youtube} dataset at 1080p resolution and 60 fps with 3 separate versions segmented at 2000/3000/4000 frames to demonstrate our model's capability for long videos. We use the standard PSNR (in dB) to measure the reconstruction quality and bits-per-pixel~(BPP) to measure the compression rate. We also include encoding times as well as their decoding speed in fps. 

The MLP network consists of 5 SIREN layers with a layer size of 512 and sine activation. The network predicts $32\times32$ patches for 3 frames ($G = 3$) in all our experiments unless mentioned otherwise. The number of iterations is set to 16000 for the first group in order to obtain a good initialization and 2000 iterations for subsequent groups. We set the learning rate to be \num{5e-4} and optimize the network with the MSE and entropy regularization loss. The entropy loss weight coefficient $\lambda_I$ as defined in Equation \ref{eqn:mse} is set to \num{1e-4}. In practice, the coefficient can be varied to control the trade-off between PSNR and BPP. We use the \textsf{torchac} library to perform arithmetic encoding of the quantized weight residuals. Since the convolutional layers typically contain only a fraction of the total network parameters (${\sim}10\%$), we do not quantize their weights and use the LZMA compression method for storing their residuals.

We use pixel-wise method SIREN \cite{sitzmann2020implicit} and frame-wise method NeRV \cite{chen2021nerv} as our baselines. For SIREN, we use a 5-layer MLP with hidden dimension of 2048. For NeRV, we use the NeRV-L configuration as specified in the paper. 
Encoding times reported are equivalent to when fully run on a single NVIDIA RTX 2080 GPU. For NeRV, we fit separate models to each video and remove 40\% of the parameters during the pruning stage with the remaining weights quantized to lowest possible bit-width without significant performance drop. Further implementation details can be found in the Appendix.

\subsection{UVG-HD}
Comparisons on the UVG-HD dataset are summarized in Table \ref{tab:baseline_comparison}. By varying the patch size, we let \ours achieve similar BPP to SIREN and NeRV respectively. \ours outperforms SIREN by a significant margin in terms of PSNR while having $6\times$ shorter encoding times. Similarly, our approach obtains speedups of ${\sim}12\times$ compared to NeRV while still achieving marginally higher PSNR (+0.34dB) and lower BPP. Additionally, we obtain a decoding speed of ${\sim}65$ FPS which is nearly $65{\times}$ and $6{\times}$ speedup in inference time/decoding compared to SIREN and NeRV respectively. This shows the efficacy of our framework to reduce redundant computation in both the spatial and temporal domains. We show our qualitative results in \Fig~\ref{fig:quality}.

\renewcommand{\arraystretch}{1.1}
\begin{table}[t]
\centering
\resizebox{\linewidth}{!}{
\begin{tabular}{@{}L{\dimexpr.15\linewidth}L{\dimexpr.33\linewidth}C{\dimexpr.27\linewidth}C{\dimexpr.25\linewidth}C{\dimexpr.13\linewidth}C{\dimexpr.1\linewidth}@{}}
\toprule
Dataset & Method & Encoding Time (Hours) $\downarrow$ & Decoding Speed (FPS) $\uparrow$ & PSNR $\uparrow$ & BPP $\downarrow$ \\
\midrule
\multirow{4}{*}{UVG-HD} & SIREN  & $\sim$30 & 15.62 &27.20 
&\textbf{0.28}\\
& \textbf{\ours (Ours)} & \textbf{5.44} & \textbf{87.91} & \textbf{34.71} & 0.32 \\
\cmidrule(l){2-6}
& NeRV  & ${\sim}$80 & 11.01 &37.36 
&0.92\\
& \textbf{\ours (Ours)}  & \textbf{6.71}&  \textbf{65.42} & \textbf{37.70} 
& \textbf{0.86}\\
\midrule
\multirow{2}{*}{UVG-4K} & NeRV & ${\sim}$134 & 8.27 & \textbf{35.24} 
& 0.28\\
& \textbf{\ours (Ours)}  & \textbf{20.89} & \textbf{50.83} & 35.18 
& \textbf{0.27}\\

\bottomrule
\end{tabular}
} 
\caption{\textbf{Comparison with video INR approaches on UVG benchmarks.}
We vary patch size of \ours on UVG-HD to match the BPP of SIREN and NeRV respectively. \ours achieved much faster encoding and decoding speed, while maintaining better or on-par quality at comparable BPP.
}
\vspace{-0.2in}
\label{tab:baseline_comparison}
\end{table}

\subsection{UVG-4K}
To analyze the spatial adaptability of our approach, we test our method on the UVG-4K dataset, with results shown in Table \ref{tab:baseline_comparison}. Compared to NeRV, we achieve a ${\sim}6\times$ speedup in both encoding and decoding times, while maintaining similar PSNR ($35.24$ vs $35.18$) and slightly better BPP ($0.28$ vs $0.27$). Furthermore, to adapt to such higher resolution data, our method does not require any architectural modifications. In contrast, NeRV requires architectural modifications by adding a $2{\times}$ upsampling block to increase the resolution of the predicted image frames. Note that a higher PSNR can be achieved with longer training schedules as we show in \Sec\ref{ssec:ablation_iter}, but we limit to 2000 iterations to maintain consistency across datasets.

\vspace{-0.5mm}
\subsection{Long Videos}
We now analyze the effect of increasing video duration for our approach. We utilize a video from the Youtube-8M dataset and evaluate on 3 separate segments consisting of the first 2000/3000/4000 frames. Results are summarized in Table \ref{tab:long_vid}. Our approach maintains a similar PSNR ($<0.3$ drop) with increased number of frames while still encoding at a similar bitrate ($<0.04$ increase). In contrast, even with higher encoding times ($4\times$ slower), NeRV suffers from significant degradation on longer videos with PSNR dropping from $33.38\rightarrow31.6\rightarrow30.53$. Since NeRV's model size remains constant, its BPP reduces with increased number of frames. However, the fixed number of network parameters limits its ability to fit to a larger set of frames, leading to performance drops. 

Additionally, since our approach is autoregressive, it needs to be trained only once even with increasing number of frames. Networks for future frames are simply initialized with the weights of the previous networks and trained before encoding the weight residuals. Such a modeling makes it suitable for online scenarios as well with a constant stream of frames. In contrast, NeRV requires separate models to be trained for different video durations as each training epoch consists of training on all frames. Note that both approaches scale linearly with increased video duration, but NeRV fits the same model to larger video signals, leading to performance drops. Specific architectural modifications are necessary to improve the PSNR for longer videos which comes at the cost of even higher encoding times.
\renewcommand{\arraystretch}{1.1}
\begin{table}[t]
\centering
\resizebox{\linewidth}{!}{
\begin{tabular}{@{}C{\dimexpr.15\linewidth}L{\dimexpr.33\linewidth}C{\dimexpr.26\linewidth}C{\dimexpr.14\linewidth}C{\dimexpr.11\linewidth}@{}}
\toprule
Num Frames & Method & Encoding Time (Hours) $\downarrow$ & PSNR $\uparrow$ & BPP $\downarrow$  \\
\midrule
\multirow{2}{*}{2000}& NeRV  & 84.44 & 33.38&0.22\\
& \ours (Ours)  & 20.85& 35.43 & 0.62\\
\midrule
\multirow{2}{*}{3000}& NeRV  & 134.58 & 31.6 &0.16\\
& \ours (Ours)  & 31.37 & 35.21 & 0.64\\
\midrule
\multirow{2}{*}{4000}& NeRV  & 190.30 & 30.53 &0.12\\
& \ours (Ours)  & 41.84 & 35.15 & 0.65\\

\bottomrule
\end{tabular}
} 
\caption{\textbf{Video duration adaptability:} For longer videos, we maintain similar reconstruction quality (${\sim}$35 PSNR) and compression rate (${\sim}$0.62 BPP). We retain a significantly faster encoding speed than NeRV which suffers from significant degradation with increased number of frames.}
\vspace{-1em}
\label{tab:long_vid}
\end{table}
\subsection{Adaptive Compression}
\label{ssec:adaptive}
\begin{figure}[t]
    \includegraphics[width=1.0\linewidth]{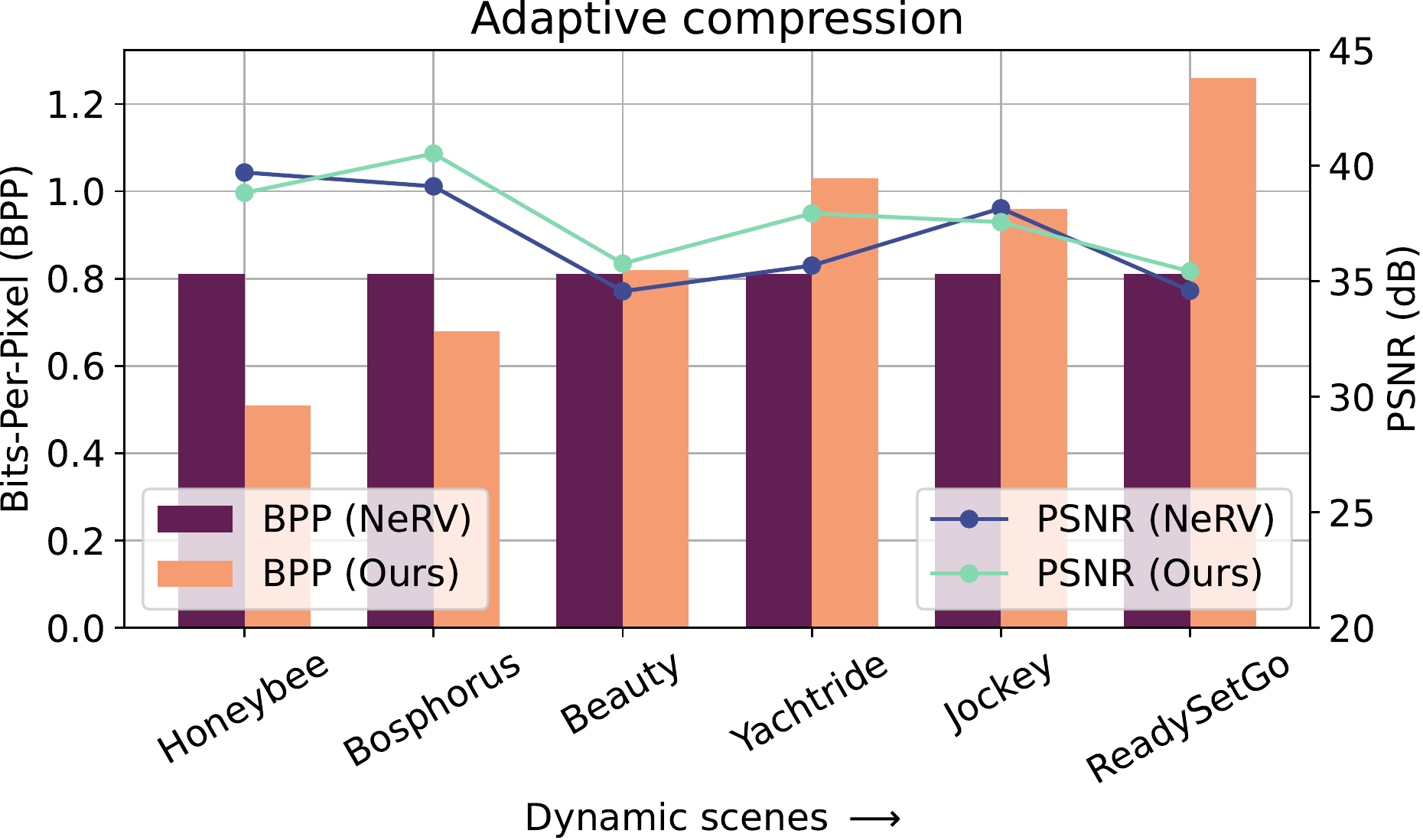}
    \caption{\textbf{Video content adaptability:} 6 videos are sorted in increasing order of variation between subsequent frames. Our approach shows adaptive bitrate compression, with more static scenes exhibiting lower BPP, while highly dynamic ones being allocated more bits while maintaining a similar PSNR as NeRV (and $12\times$ encoding speed).}
    \label{fig:adaptive}
    \vspace{-1em}
\end{figure}

Videos can consist of different levels of inter-frame motion with more static videos containing higher levels of temporal redundancy in comparison to dynamic ones. To illustrate the capability of our approach to exploit such redundancies, we evaluate the compression rate on 6 videos in the UVG-HD dataset which consist of 600 frames. We sort each video according to their average MSE between subsequent frames, which serves as a proxy to the amount of temporal redundancy in the video. More static scenes like Honeybee have a lower MSE compared to highly dynamic scenes like Jockey. We plot the PSNR and BPP of \ours and NeRV with increasingly dynamic video content and show the results in \Fig\ref{fig:adaptive}. Note that the average PSNR/BPP over the 6 videos can be increased or decreased by varying other hyperparameters such as patch size, number of groups, entropy loss coefficient etc. (as shown in \Sec\ref{sec:ablation}), but we focus on adaptability to videos for a given hyperparameter configuration. We see that our approach has an adaptive bitrate compression with more static scenes like Honeybee (MSE \num{2.2e-4}) that has a lower bitrate (0.51), compared to dynamic scenes such as Jockey (MSE \num{0.9e-3}) which are allocated more bits (0.96). We maintain similar PSNR as NeRV which has a constant BPP due to the same model applied to all videos. While NeRV's quantization bit width can be reduced further for lower BPP, it is a post-hoc approach which comes at the cost of PSNR and requires tuning for each video. In contrast, our approach adaptively varies the BPP during training with no change in hyperparameters.

\subsection{GPU Parallelization}
\label{ssec:gpu_parallel}
\begin{figure}[t]
    \includegraphics[width=1.0\linewidth]{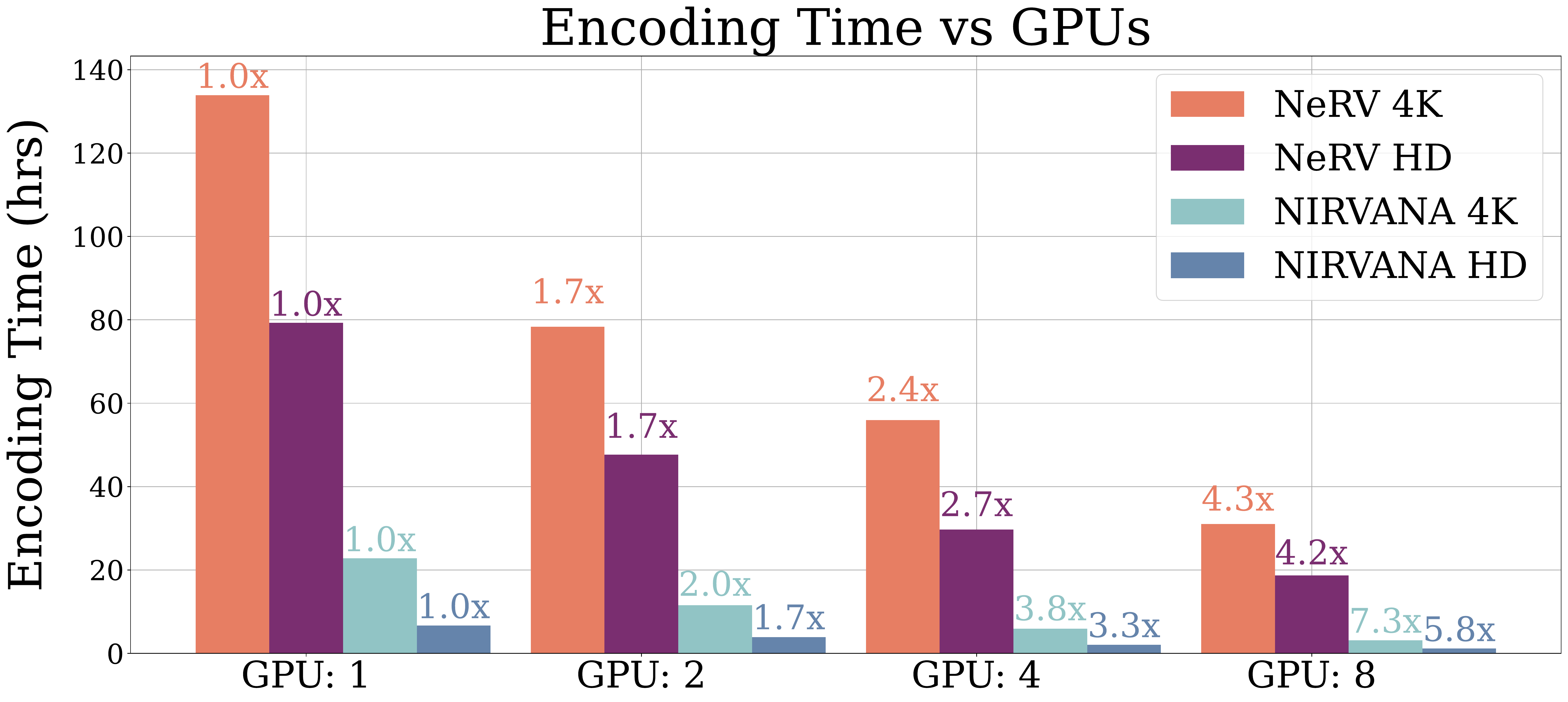}
    \caption{\textbf{GPU scalability of \ours:} We compare scalability of our approach with NeRV in terms of encoding time with increasing number of GPUs at two video resolutions: 1080p and 4K. We scale close to linear for 4K and have much lower overhead compared to NeRV for both resolutions.}
    \label{fig:gpu}
    \vspace{-0.1in}
\end{figure}

We now analyze the scalability of our approach with a larger number of GPUs. In \Fig~\ref{fig:gpu} we plot the encoding times for ``Jockey" (both 1080p and 4K versions) from UVG dataset for NeRV (using distributed training) and our methods. The design in Algorithm \ref{alg:vid_INR} allows different \textit{chunks} of the source video to be processed autoregressively on separate GPUs. As the number of GPUs are scaled with a factor of $2\times$, our approach achieves close to linear scaling with very little overhead for the case of UVG-4K ($1.0\times\rightarrow2.0\times\rightarrow3.8\times\rightarrow7.3\times$) compared to a weaker scaling of NeRV ($1.0\times\rightarrow1.7\times\rightarrow2.7\times\rightarrow4.3\times$). UVG-HD shows a higher amount of overhead but still scales fairly well with increased GPUs compared to NeRV. Thus, we see the capability of parallelization of our approach with higher number of GPUs. Also note that the time taken by NeRV for HD videos on 8 GPUs is still higher than the time taken by \ours on a single GPU. 

\section{Ablation Studies}
\label{sec:ablation}
In this section, we study the impact of various parameters of our approach on the PSNR-BPP tradeoff curves as seen in \Fig\ref{fig:ablation}. By varying the entropy loss coefficient, we obtain different points on the tradeoff curve with a higher coefficient leading to lower BPP (low rate) but also lower PSNR (high distortion). We additionally show the convergence effects of longer training for each group with increased number of iterations. Results are evaluated on the Jockey video of the UVG-HD dataset. While varying each parameter along with the entropy coefficient, we fix other parameters to their default values of patch size at $32\times32$, group size at 3, and number of iterations at 2000. We sample values of the entropy coefficient $\lambda_I$ between \num{1e-5} and \num{5e-4} to obtain various points on the tradeoff curve.
\subsection{Effect of Entropy Regularization}
\label{ssec:ablation_entropy}
\begin{figure*}[t]
    \centering
    \setlength{\tabcolsep}{1pt}
    \begin{tabular}{cccc}
    \includegraphics[width=0.25\linewidth]{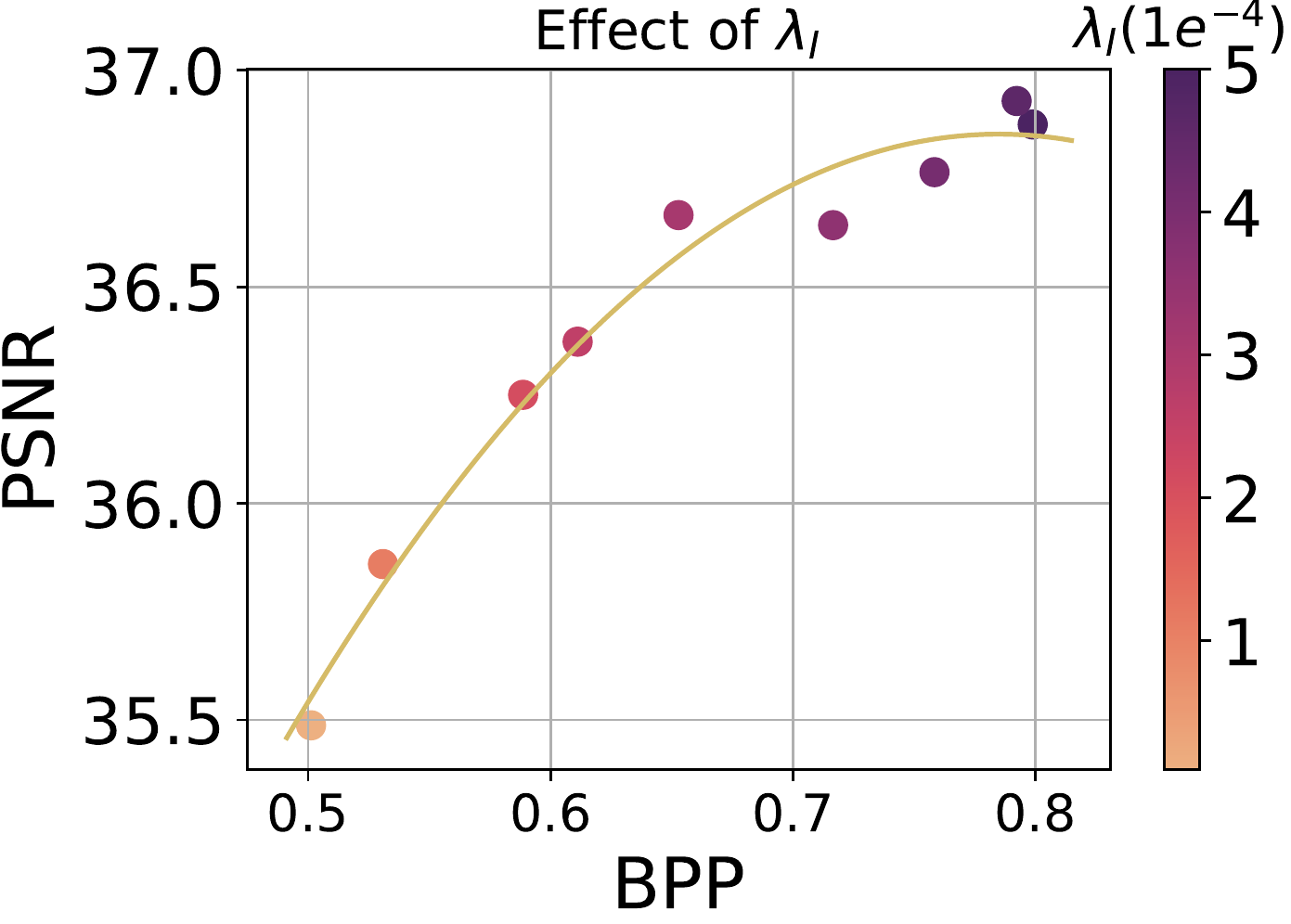}&
    \includegraphics[width=0.24\linewidth]{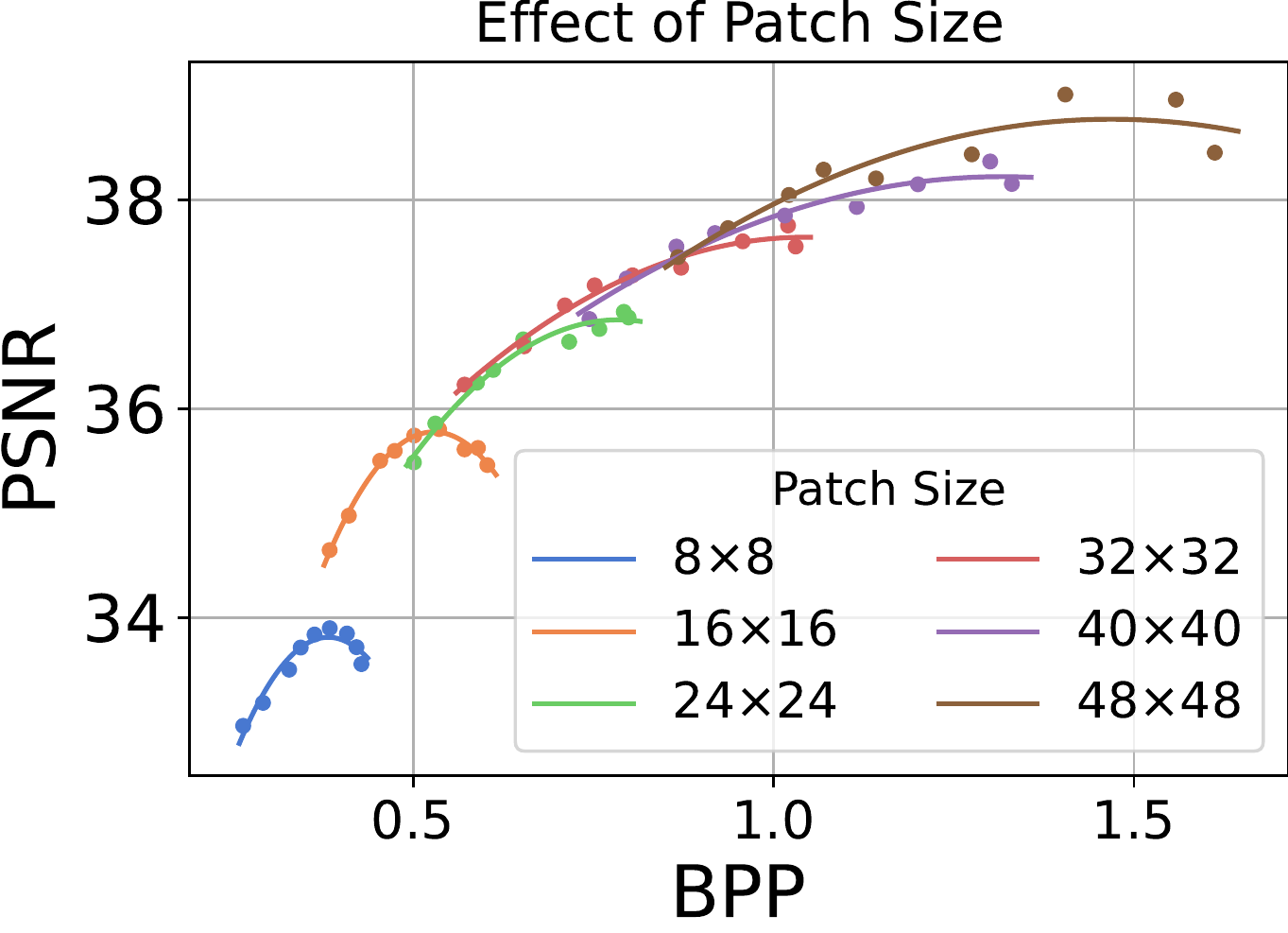}&
    \includegraphics[width=0.24\linewidth]{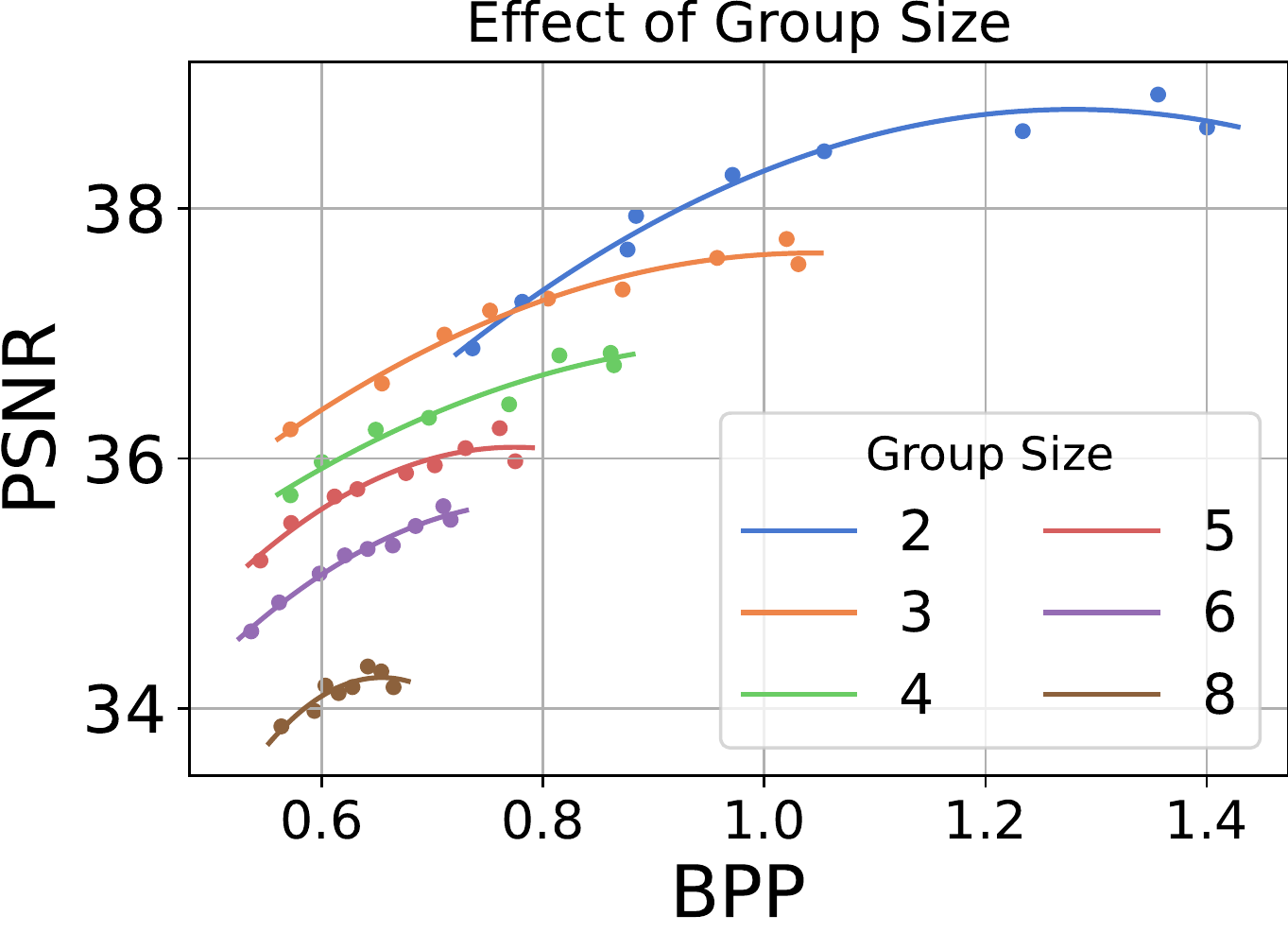}&
    \includegraphics[width=0.24\linewidth]{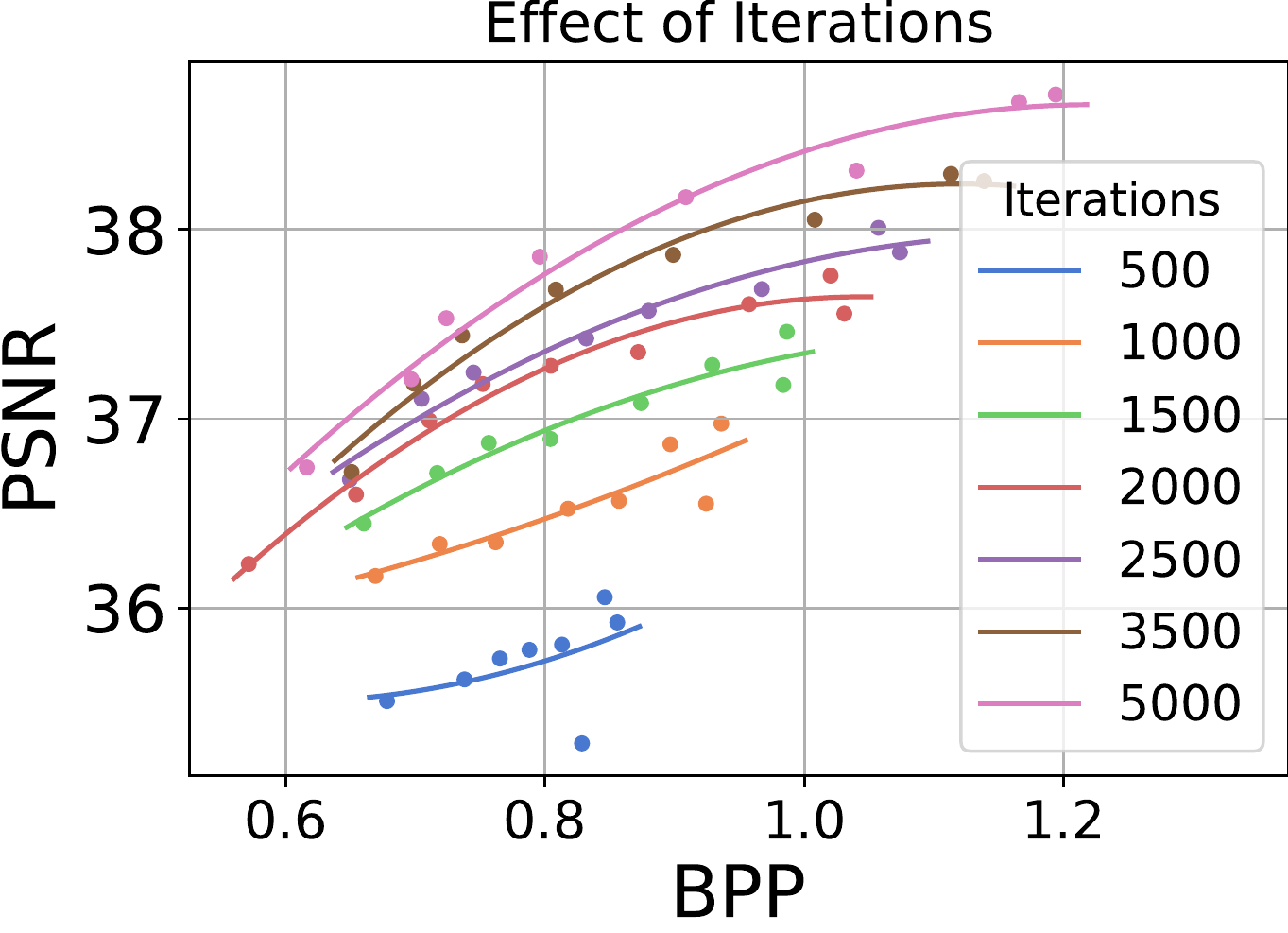}\\
    (a)&\quad\enskip(b)&\quad\enskip (c)&\qquad(d)
    \end{tabular}
    \vspace{-0.15in}
    \caption{(a) Effect of entropy regularization: the larger $\lambda_I$ is, the lower the entropy and the BPP. (b, c) As the patch size or group size increases, PSNR increases at the cost of higher BPP. (d) The longer the training iteration (encoding time) is, the higher the PNSR gains. }
    \vspace{-0.1in}
    \label{fig:ablation}
\end{figure*}

We analyze the effect of varying the entropy coefficient $\lambda_I$ and obtain a PSNR-BPP curve visualized in \Fig\ref{fig:ablation}(a). In general, we see that increasing $\lambda_I$ decreases the BPP at the cost of PSNR. This is to be expected as increasing the entropy regularization forces the quantized weights of each frame group's networks to lie in fewer quantization bins. Consequently, more weight residual (difference between quantized weights of subsequent frame groups' networks) values are 0 leading to lower entropy of the weights and subsequently lower BPP of the model. The entropy coefficient thus provides a natural way of controlling the PSNR-BPP tradeoff according to the required application.

\subsection{Effect of Patch Size}
\label{ssec:ablation_patch}
We vary the patch prediction size of our network from $8\times8$ to $48\times48$ in steps of 8. We visualize the results in \Fig\ref{fig:ablation}(b). In general, increasing patch size shifts the curve upwards and to the right corresponding to higher PSNR but also high BPP. A higher patch size results in an increase in number of network parameters (both in convolutional and SIREN layers) and hence its expressivity, leading to higher PSNR. However, as patches are less localized, the outputs of networks between subsequent frame-groups vary more significantly with dynamic scenes (such as Jockey), leading to larger residuals. This increases the entropy of the residuals and as a result, the BPP.

\begin{figure*}[t]
    \centering
    \begin{subfigure}[b]{1.0\textwidth}
         \centering
        \includegraphics[width=0.33\textwidth]{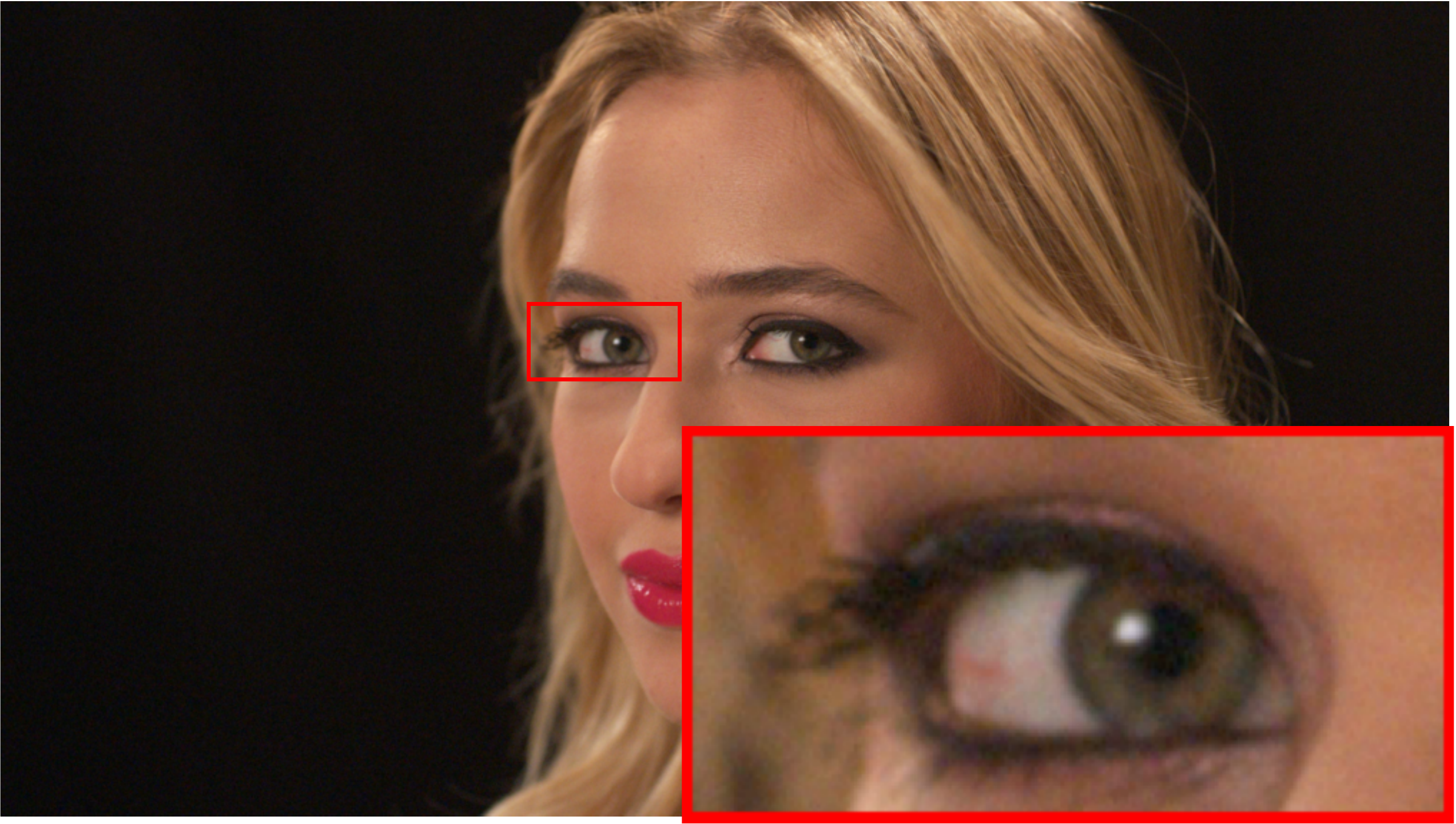}
        \includegraphics[width=0.33\textwidth]{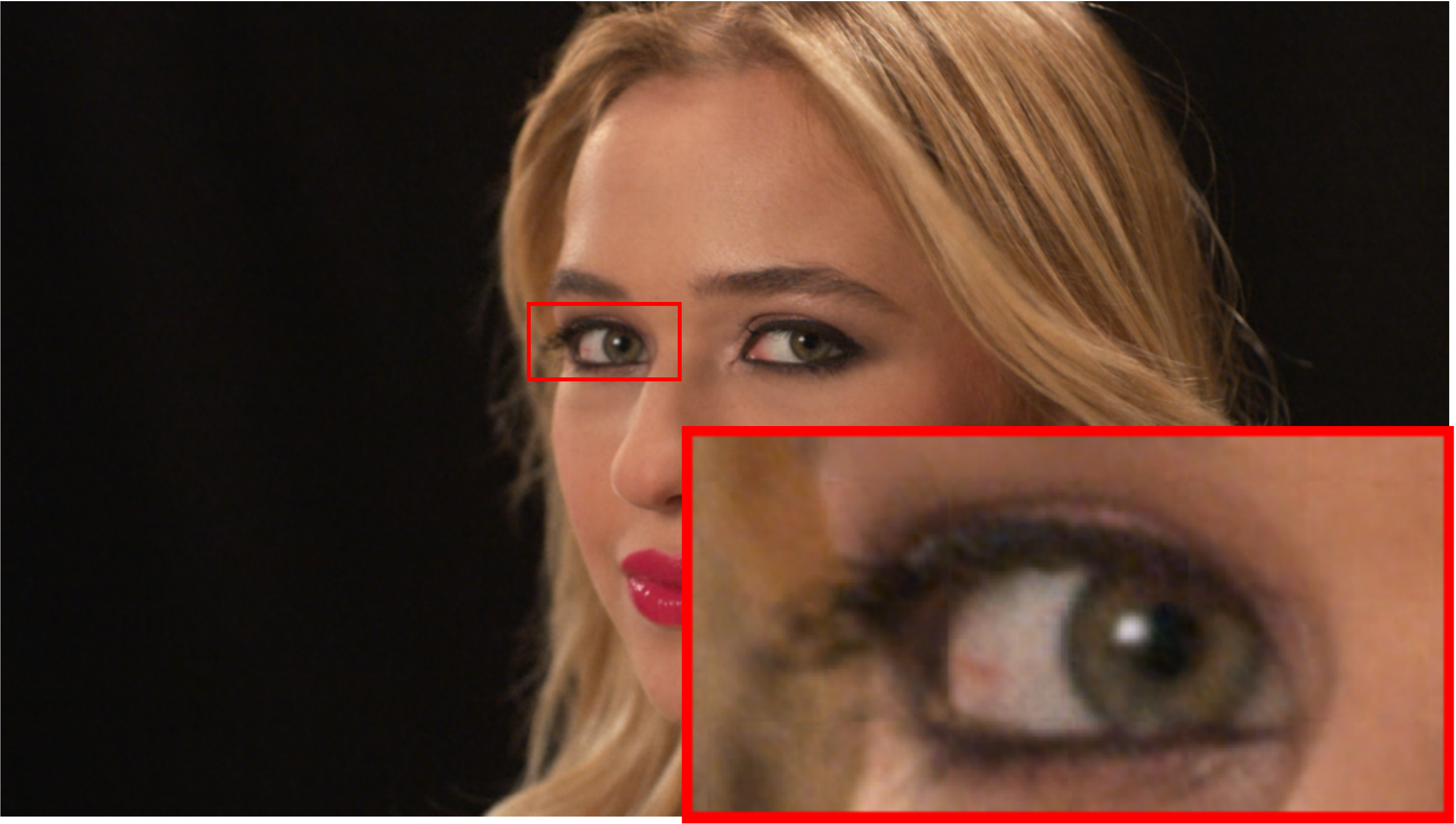}
        \includegraphics[width=0.33\textwidth]{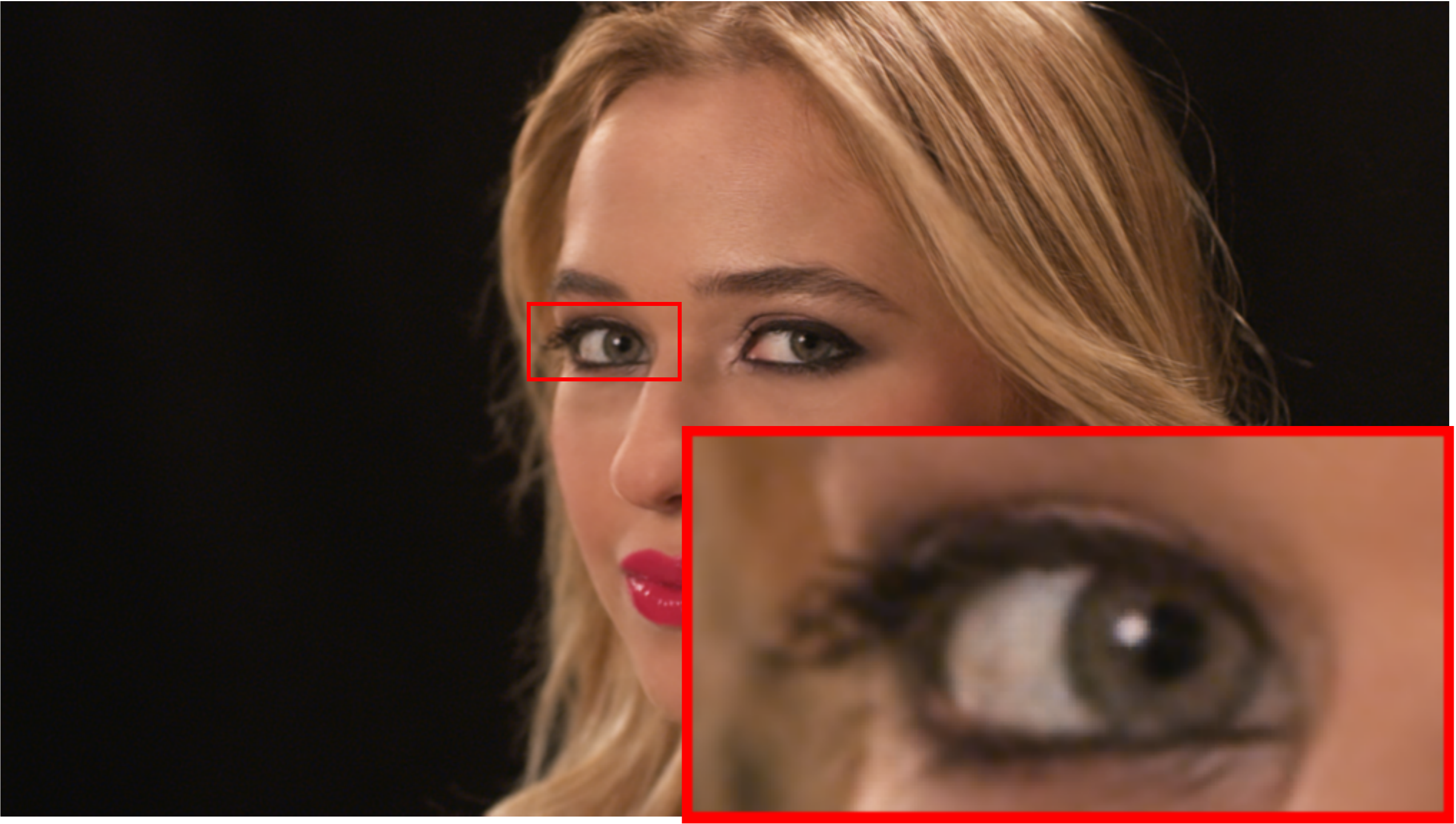}
    \end{subfigure}
    \begin{subfigure}[b]{1.0\textwidth}
         \centering
        \includegraphics[width=0.33\textwidth]{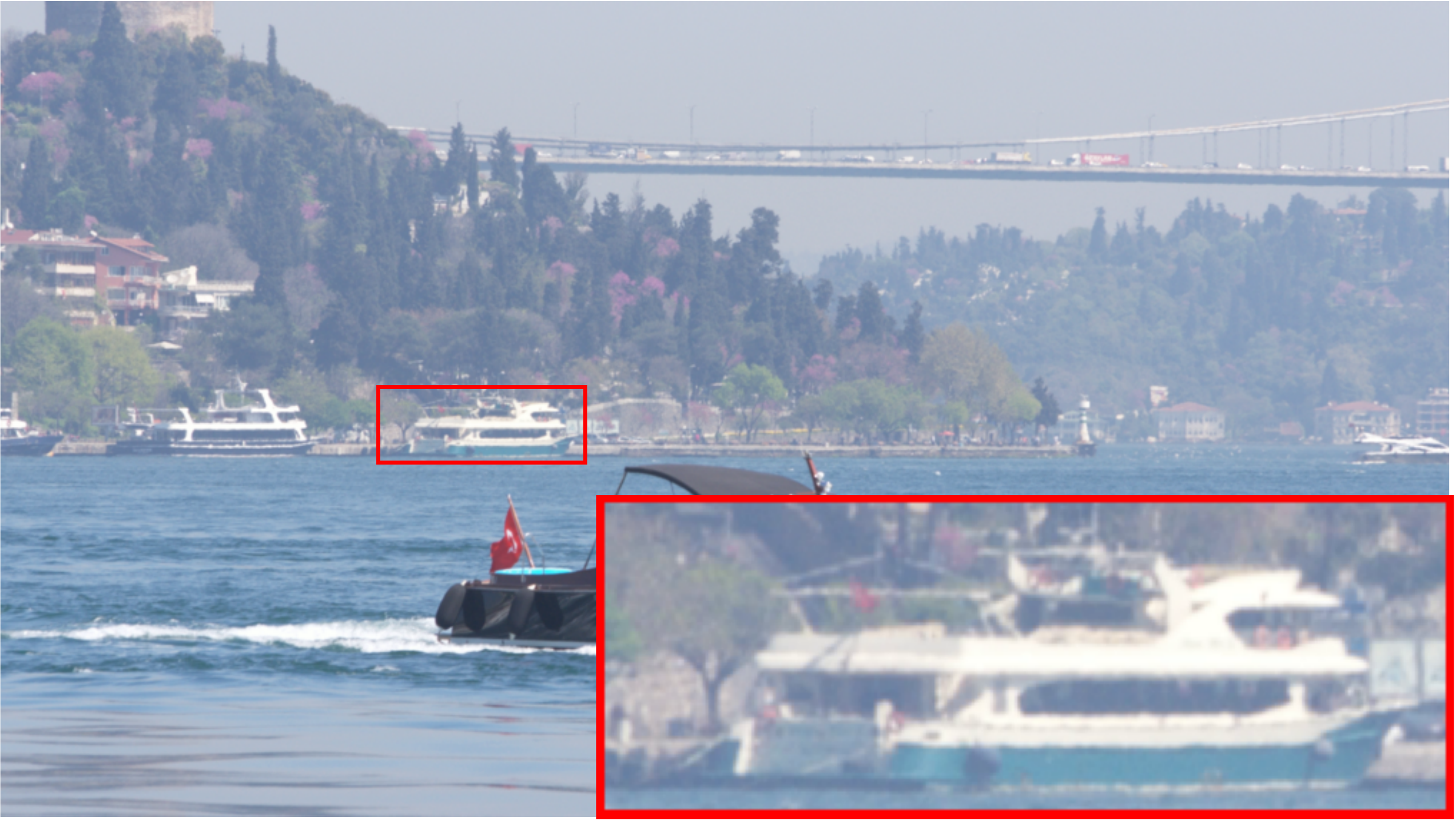}
        \includegraphics[width=0.33\textwidth]{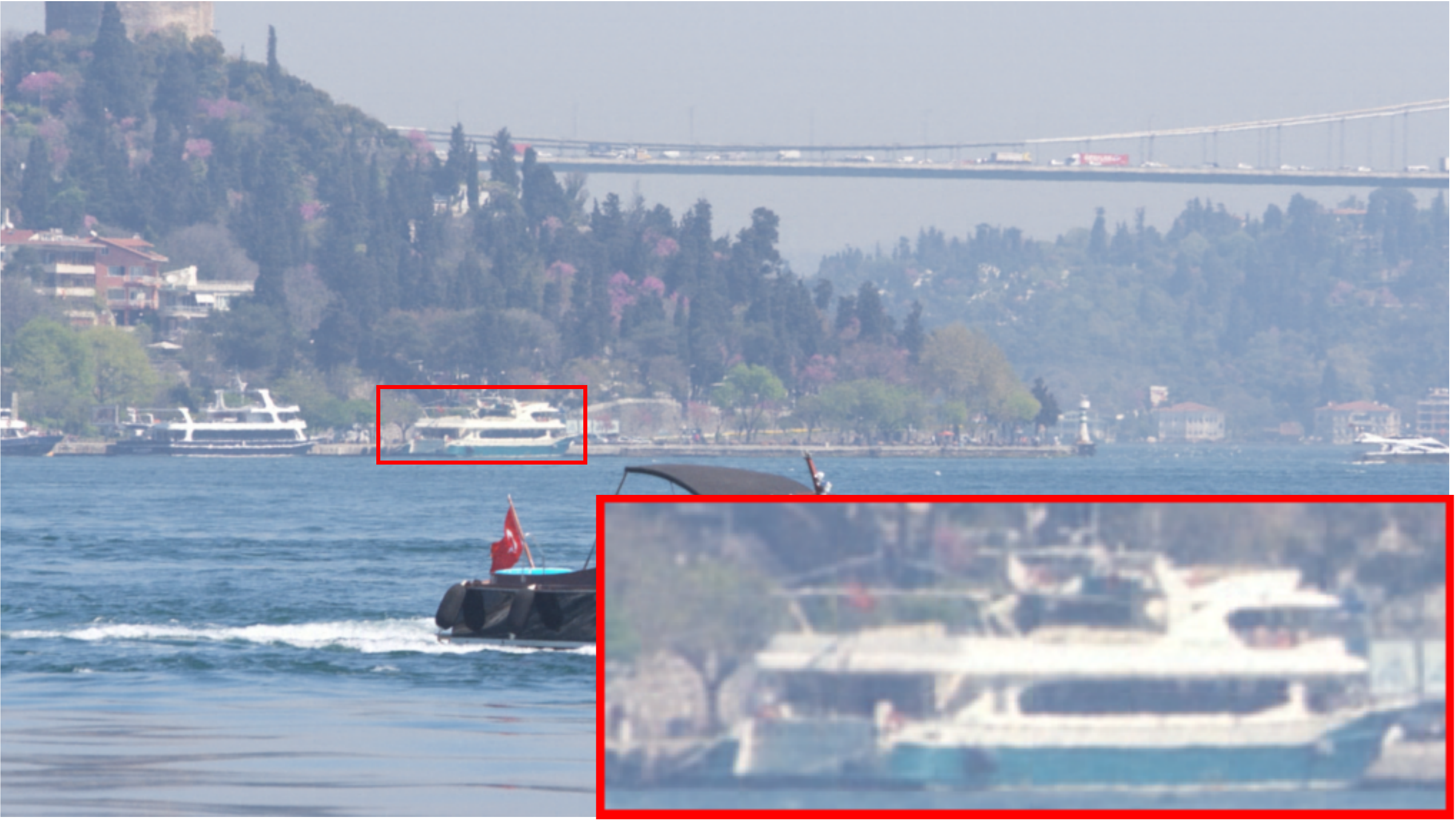}
        \includegraphics[width=0.33\textwidth]{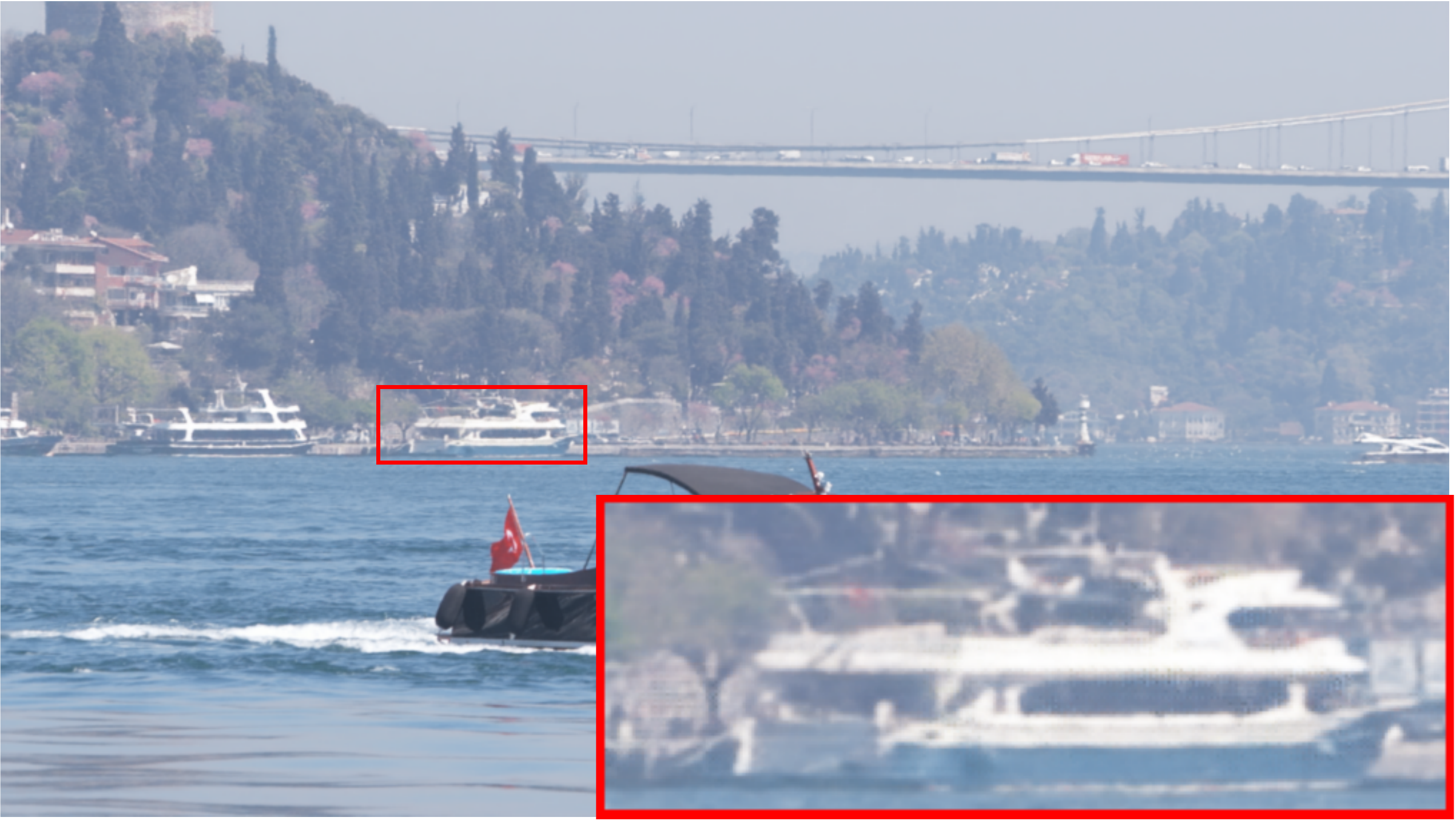}
    \end{subfigure}    
    \vspace{-0.2in}
    \caption{(Left) Ground truth video frames. (Center) Reconstruction from \ours. (Right) Reconstruction from NeRV. We show that \ours is able to preserve the image fidelity after reconstruction, capturing important details such as the veins in the eye of Beauty, and the color quality in the Bosphorus video.}
    \label{fig:quality}
    \vspace{-0.15in}
\end{figure*}
\subsection{Effect of Frame Group Size}
\label{ssec:ablation_group}
We vary the frame group size from $2$ to $8$ in steps of $1$ and visualize the results in \Fig\ref{fig:ablation}(c). Increasing the group size, in general, reduces BPP at the cost of PSNR. This is expected as a single model shares computation across a larger group of frames effectively leading to fewer parameters per frame and lower BPP. However, we notice that group size $3$ obtains the best tradeoff curve in the $<0.8$ BPP regime with higher group size detrimental to the performance. This is likely because of the fixed MLP representation capability which learns a shared global representation for all frames within a group. For a dynamic scene such as Jockey with significant pixel shift between frames (\Sec\ref{ssec:adaptive}), a larger MLP is necessary for capturing the variations within a group. 

\subsection{Effect of Number of Training Iterations}
\label{ssec:ablation_iter}
To analyze the effect of longer training schedules, we vary the number of training iterations for the network for each frame group from $500$ to $5000$. Results are shown in \Fig\ref{fig:ablation}(d). As expected, increasing the number of iterations improves the PSNR-BPP tradeoff with the curve shifting upwards. This shows that our network can obtain higher quality reconstructions for longer training times at no cost to the compression rate. This can be made feasible with higher number of GPUs as shown in \Sec\ref{ssec:gpu_parallel}. However, increasing iterations provides diminishing gains as we observe the curves approaching closer to each other with higher iterations. 
\vspace{-0.5mm}

\section{Qualitative results}
We visualize the reconstructions of frames in videos from the UVG-HD dataset in \Fig~\ref{fig:quality}. Our approach (center) preserves image fidelity after reconstruction when compared to the ground-truth frame (left). It captures subtle details such as the veins in the eye (top row) or maintaining the color information of the boat (bottom row) when compared to the ground-truth. Further visualizations from videos in the UVG-4K dataset are provided in the Appendix.
\section{Conclusion}
In this work, we propose an autoregressive video INR framework, \ours, which segments videos into groups of frames and fits separate neural networks to each group. Each network performs a patch-wise prediction across the group of frames thus exploiting both the spatial and temporal redundancy present in videos, improved from the previous works. Each network is initialized with the weights of the previous frame-group's trained network. We additionally quantize weights during training, requiring no post-hoc pruning or quantization and store weight residuals between subsequent frame group's networks to obtain high levels of compression. \ours achieves $12{\times}$ speedups on standard datasets compared to previous methods while maintaining similar levels of reconstruction quality and compression rate. \ours adapts to varying video resolution and duration without large performance degradation and no architectural modifications. Additionally, our framework adaptively compresses videos based on their inter-frame variation. We achieve high levels of decoding speed compared to prior video INR approaches and also scale better with higher number of GPUs.




{\small
\bibliographystyle{ieee_fullname}
\bibliography{main}
}

\clearpage
\appendix
\maketitle
{\LARGE\bfseries\noindent Appendix}
\section{Additional Implementation Details}

\subsection{NeRV}
We use the NeRV-L config from the original paper. The model takes in positional embedding of time coordinate as input. We set the number of sine levels to 80 and base value to 1.25. The hidden dimension of the 2-layer MLP in the beginning is set to 1024, with 128 output channels and an $8\times$ channel expansion for the first NeRV Block. We stack 5-NeRV blocks with upscale factors of $\{5,3,2,2,2\}$ for HD version and an additional block with $2\times$ upsampling for the 4K version. Following the standard training schedule, we use cosine learning rate schedule starting with $5e^{-4}$, with a warmup of 0.2. We use the combination of L1 and SSIM loss and train the model with a batch size of 1, as mentioned in the paper. 

\subsection{Entropy Loss and Weight Quantization}
We follow the works of \cite{oktay2019scalable,Girish2022LilNetXLN} for performing model compression with small variations. We represent our MLP layer weights as ${\bw_1,\bw_2,...,\bw_L}$ for L layers with the $l^{th}$ layer weight matrix $\bw_l \in \mathbb{R}^{O_l\times I_l}$ having a shape $O_l\times I_l$ consisting of continuous values. For each weight matrix, we maintain a corresponding flattened latent quantized representation vector $\bwt_l \in \mathbb{Z}^{O_lI_l}$. $\bwt_l$ consists of integer values for each corresponding element in $\bw_l$. For ease of notation, we drop the subscript $l$.

We then maintain decoders $f_\phi(.)$ parameterized by parameters $\phi$. The weight matrix $\bw$ is then obtained from the quantized weights $\bwt$ as $\bw = f_\phi(\bwt)$. Prior works use matrices or vectors to represent the weight decoders while we use a single scalar $\phi$ as a parameter of the decoder. The weight matrix is thus simply, $\bw = reshape(\phi\bwt)$, where the scale parameter $\phi$ is multiplied with individual values of $\bwt$. We make the scale parameter learnable by passing gradients. Thus, it effectively controls the bit width of the weight matrix. We maintain separate decoders for each layer of the MLP.

To make the network fully differentiable, we maintain continuous surrogates $\widehat{\bw}$ for the discrete latents $\bwt$. During training, we simply round the surrogates to their nearest integer to obtain the discrete latents which are then passed to the decoder. We make the rounding operation differentiable using the straight-through estimator \cite{bengio2013estimating} to pass the gradients from $\widehat{\bw}$ to $\bwt$. 

To reduce the entropy of the quantized latents, we use probability models from \cite{balle2018variational}. For each continuous surrogate $\widehat{\bw}$, we maintain probability models $c_\theta(.)$ parameterized by $\theta$ which output the CDF of the latent distributions. Similar to prior works, we use uniform noise $n\sim\mathcal{U}\left(-\frac{1}{2},\frac{1}{2}\right)$ as a substitute for quantization. The entropy of the model can now be minimized by minimizing the self-information $\boldsymbol{I}$ as follows:
\begin{align}
    I(\bwt) = -\log_2(c_\theta(\widehat{\bw}+n)).
\end{align}
This serves as the entropy regularization loss which controls the rate-distortion tradeoff. A higher entropy coefficient $\lambda_I$ leads to more compressed latents (lower rate), but usually at the cost of PSNR (higher distortion).
The network latents, decoder parameter, and probability model parameters are learnable and jointly optimized. Following prior works, we use a learning rate of $1e{-}4$ for the probability model weights and the same learning rate for the decoder weights. We set the learning rate of the latents to $5e{-}4$.  All the parameters are optimized in an end-to-end manner with an Adam optimizer during training, thus requiring no post-hoc approaches. After training, we discard the probability models and use the frequency of each quantized value in the latent vector to obtain the probability tables required for arithmetic entropy coding. Note that the continuous surrogates are discarded and only their rounded discrete latents are stored using entropy coding. These latents can then be decoded using the probability tables. The decoder parameters and probability tables have almost no overhead compared to the overall model latents.

\begin{table}[t]
\resizebox{\linewidth}{!}{
\begin{tabular}
{@{}L{\dimexpr.26\linewidth}C{\dimexpr.1\linewidth}C{\dimexpr0.6\linewidth}}
\toprule
Video & Frames & Link \\
\midrule
Mario Kart & 4000 & \url{http://bit.ly/3XjIvfR} \\
Dota & 4261 & \url{http://bit.ly/3Xf6Nru} \\
Ride & 4000 & \url{http://bit.ly/3TOEgWI} \\
Submarine & 3626 & \url{http://bit.ly/3EJKzGM}\\
Water Scooter & 4199 & \url{http://bit.ly/3OhF99h}\\
Mortal Kombat & 3239 & \url{http://bit.ly/3EdvNGU}\\
\bottomrule
\end{tabular}
}
\caption{\textbf{Details of videos from the Youtube-8M dataset.}}
\label{youtube8m}
\vspace{-0.6cm}
\end{table}

\section{Additional Dataset Details} 
\subsection{UVG-4K}
In addition to the datasets shown in Section 4 of the main paper, we show quantitative and qualitative results on 7 more videos from the UVG dataset at the 4K resolution: Twilight, Sunbath, CityAlley, FlowerFocus, FlowerKids, RiverBank, and RaceNight. We dub this dataset UVG-4K (Set 2). 

\subsection{Youtube-8M}

We select 5 more videos from the Youtube-8M dataset, with varying video content to further test the ability of our model to encode longer videos. This is an extension of the experiments from Section 4.4 which consists of a single video (Mario Kart).
We present the details of each video used in Table~\ref{youtube8m}.

\renewcommand{\arraystretch}{1.1}
\begin{table*}[t]
\centering
\resizebox{0.7\linewidth}{!}{
\begin{tabular}{@{}lcccccccc@{}}
\toprule
\multirow{2}{*}{Video Name} & \multicolumn{4}{c}{\ours (Ours)}& \multicolumn{4}{c}{NeRV}\\
\cmidrule(l){2-5}\cmidrule(l){6-9}
&PSNR $\uparrow$&VMAF $\uparrow$&FLIP $\downarrow$&BPP $\downarrow$&PSNR $\uparrow$&VMAF $\uparrow$&FLIP $\downarrow$&BPP $\downarrow$\\
\midrule
ReadySteadyGo & \textbf{35.43} &\textbf{98.04} &\textbf{0.0862} &1.26 &34.59 &96.95 &\textbf{0.0862} &\textbf{0.81}\\
Bosphorus  & \textbf{40.53} &\textbf{90.97} &\textbf{0.0549} &\textbf{0.68} &39.11 &88.26 &0.0621 &0.81\\
Beauty     & \textbf{35.77} &84.46 &\textbf{0.0524} &0.96 &34.57 &\textbf{86.35} &0.0605 &\textbf{0.81}\\
Honeybee   & 38.83 &91.31 &0.0505 &\textbf{0.51} &\textbf{39.71} &\textbf{95.08} &\textbf{0.0419} &0.81\\
Jockey     & 37.56 &93.07 &0.0710 &0.96 &\textbf{38.16} &\textbf{95.76} &\textbf{0.0653} &\textbf{0.81}\\
Yachtride  & \textbf{37.94} &\textbf{91.31} &\textbf{0.0668} &1.03 &35.68 &88.70 &0.0786 &\textbf{0.81}\\
ShakeNDry  & 37.82 &88.81 &0.0608 &\textbf{0.76} &\textbf{39.68} &\textbf{95.20} &\textbf{0.0468} &1.61\\
\midrule
Average    & \textbf{37.70} &91.14 &0.0632 &\textbf{0.86} &37.35 &\textbf{92.33} &\textbf{0.0631} &0.92\\
\bottomrule
\end{tabular}
} 
\caption{\textbf{Video-wise performance on UVG-HD:} We show video-wise results of the 7 videos in UVG-HD and compare the reconstruction quality using the 3 metrics of PSNR, VMAF, FLIP along with compression rate measured by BPP. We see that we maintain similar performance as NeRV in all 3 metrics and BPP while having 12$\times$ faster encoding speed (as also shown in Table 1 of the main paper).
}
\label{tab_supp:uvg_hd_videowise}
\end{table*}

\renewcommand{\arraystretch}{1.1}
\begin{table*}
\centering
\resizebox{\linewidth}{!}{
\begin{tabular}{@{}C{\dimexpr.4\linewidth}C{\dimexpr.4\linewidth}@{}}
\resizebox{1.0\linewidth}{!}{
\begin{tabular}{@{}L{\dimexpr.35\linewidth}C{\dimexpr.3\linewidth}C{\dimexpr.15\linewidth}C{\dimexpr.15\linewidth}C{\dimexpr.15\linewidth}@{}}
\toprule
\multirow{2}{*}{Video Name} & \multicolumn{2}{c}{\ours (Ours)}& \multicolumn{2}{c}{NeRV}\\
\cmidrule(l){2-3}\cmidrule(l){4-5}
&PSNR&BPP&PSNR&BPP\\
\midrule
ReadySteadyGo & \textbf{33.85}& 0.41&33.22 &\textbf{0.24}\\
Bosphorus &38.71&\textbf{0.21}&\textbf{39.0} &0.24\\
Beauty &\textbf{31.96}&0.28&31.05 &\textbf{0.24}\\
Honeybee &35.64&\textbf{0.14}&\textbf{36.36} &0.24\\
Jockey &35.05&0.30&\textbf{35.9} &\textbf{0.24}\\
Yachtride &\textbf{36.33}&0.33&35.05 &\textbf{0.24}\\
ShakeNDry &34.78&\textbf{0.24}&\textbf{36.09} &0.49\\
\midrule
Average &35.18&\textbf{0.27}&\textbf{35.23} &0.28\\
\bottomrule
\end{tabular}
}&
\resizebox{1.0\linewidth}{!}{
\begin{tabular}{@{}L{\dimexpr.35\linewidth}C{\dimexpr.3\linewidth}C{\dimexpr.15\linewidth}C{\dimexpr.15\linewidth}C{\dimexpr.15\linewidth}@{}}
\toprule
\multirow{2}{*}{Video Name} & \multicolumn{2}{c}{\ours (Ours)}& \multicolumn{2}{c}{NeRV}\\
\cmidrule(l){2-3}\cmidrule(l){4-5}
&PSNR&BPP&PSNR&BPP\\
\midrule
FlowerFocus & 36.50& \textbf{0.12}&\textbf{37.08} &0.24\\
CityAlley& 37.43 & \textbf{0.17} &\textbf{38.39} &0.24\\
Twilight & \textbf{38.02} & \textbf{0.13} &20.99 &0.24\\
FlowerKids & \textbf{34.62}& 0.26&33.77 &\textbf{0.24}\\
RiverBank &\textbf{33.83}&0.26&32.35 &\textbf{0.24}\\
RaceNight &32.72&0.27&\textbf{32.92} &\textbf{0.24}\\
Sunbath &37.75&\textbf{0.24}&\textbf{44.17} &0.49\\
\midrule
Average &\textbf{35.84}&\textbf{0.21}&34.23 &0.28\\
\bottomrule
\end{tabular}
}\\
\quad\resizebox{0.3\linewidth}{!}{(a) UVG-4K (Set 1)} & \quad\resizebox{0.3\linewidth}{!}{(b) UVG-4K (Set 2)}

\end{tabular}
} 

\vspace{-0.2in}
\caption{\textbf{Video-wise comparison on different sets of UVG-4K:} We show video-wise results on 2 different sets of 7 UVG videos at 4K resolution. Set-1 consists of videos from UVG-HD at 4K resolution while Set 2 consists of additional 7 videos from the dataset. We maintain similar performance in terms of PSNR and BPP as NeRV while also being ${\sim}6\times$ faster for both sets and being $6\times$ faster in terms of encoding time.
}
\label{tab_supp:uvg_4k_videowise}
\end{table*}

\section{Video-wise comparison}

We show additional quantitative results on UVG-4K (Set 2) and Youtube-8M mentioned above.

\subsection{UVG-HD}
We provide video-wise results of our approach along with that of NeRV \cite{chen2021nerv}. We evaluate the video on the additional perceptual quality metrics of FLIP \cite{10.1145/3406183} and VMAF \cite{vmaf} as well, along with the standard PSNR. Results are summarized in Table \ref{tab_supp:uvg_hd_videowise}. We see that we continue to obtain similar performance compared to NeRV in terms of these metrics while being ${\sim}12\times$ faster. Also note the adaptive BPP of our method, which is based on the amount of motion in each video. In contrast, NeRV maintains a fixed BPP due to fixed model size (ShakeNDry shows twice the BPP due to half the number of frames). We observe a small drop in VMAF ($92.33\rightarrow91.14$) while maintaining similar value of FLIP ($\sim0.0632$) compared to NeRV.

\subsection{UVG-4K}
We provide video-wise results on the 2 sets of UVG at 4K resolution. For Set 1, we obtain comparable performance to NeRV while obtaining ${\sim}6\times$ faster encoding speed. Similar to UVG-HD, we continue to show the benefits of adaptive compression, with static videos such as Honeybee showing lower levels of BPP (0.14) compared to the most dynamic video, ReadySteadyGo (0.41 BPP). For Set 2, we outperform NeRV by $1.5$ PSNR while still obtaining $25\%$ lower BPP $0.28\rightarrow0.21$. 
The PSNR drop of NeRV on the Twilight video is largely due to quantization at the fixed bit width of 20.
Hand-tuning is necessary in order to maintain higher PSNR but at the cost of BPP. In contrast, our approach maintains the reconstruction quality for a variety of videos and adaptively quantizes for each video.

\begin{figure*}[t]
    \centering
    \setlength{\tabcolsep}{1pt}
    \begin{tabular}{cc}
    \includegraphics[width=0.45\linewidth]{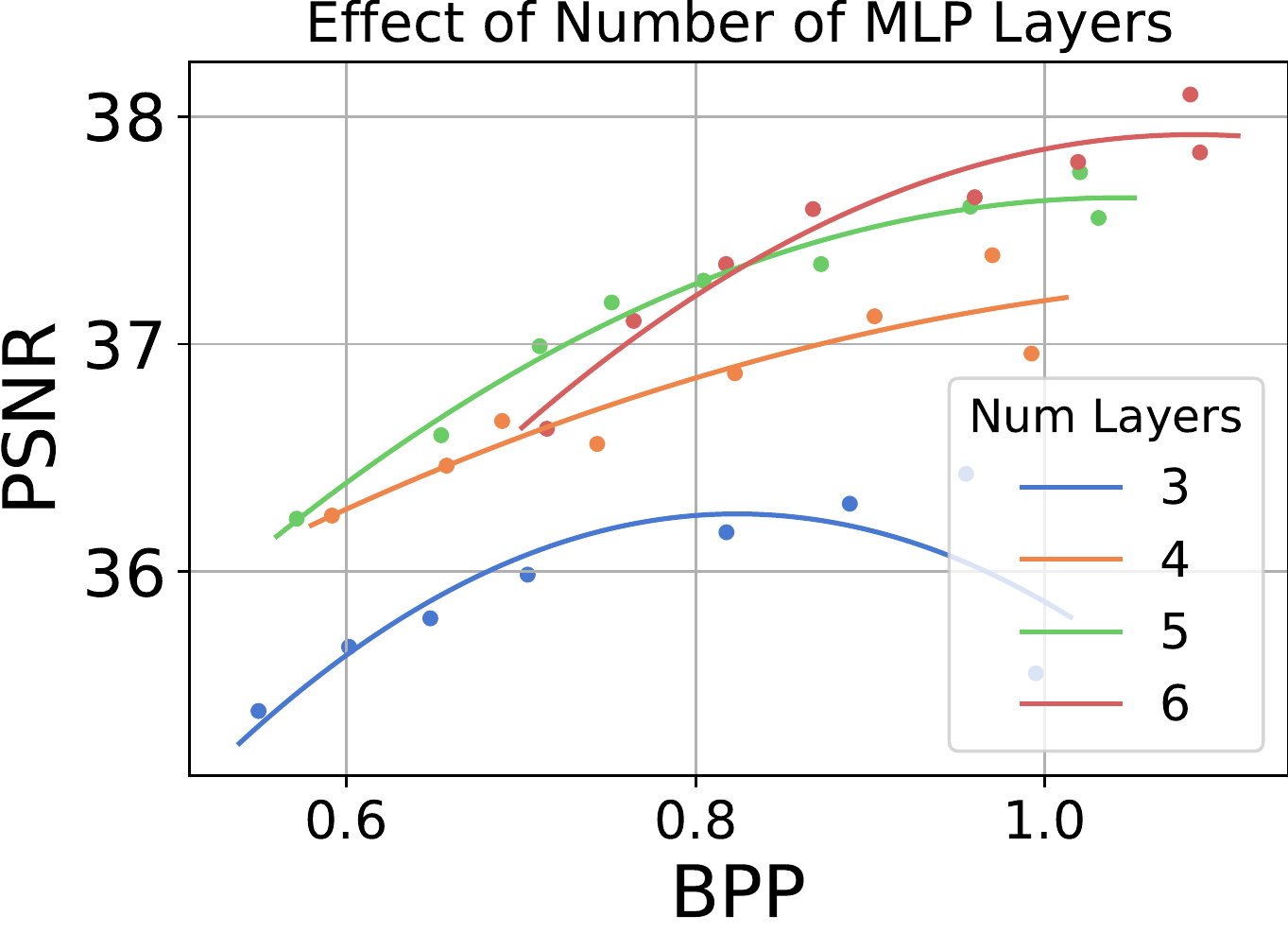}&
    \includegraphics[width=0.44\linewidth]{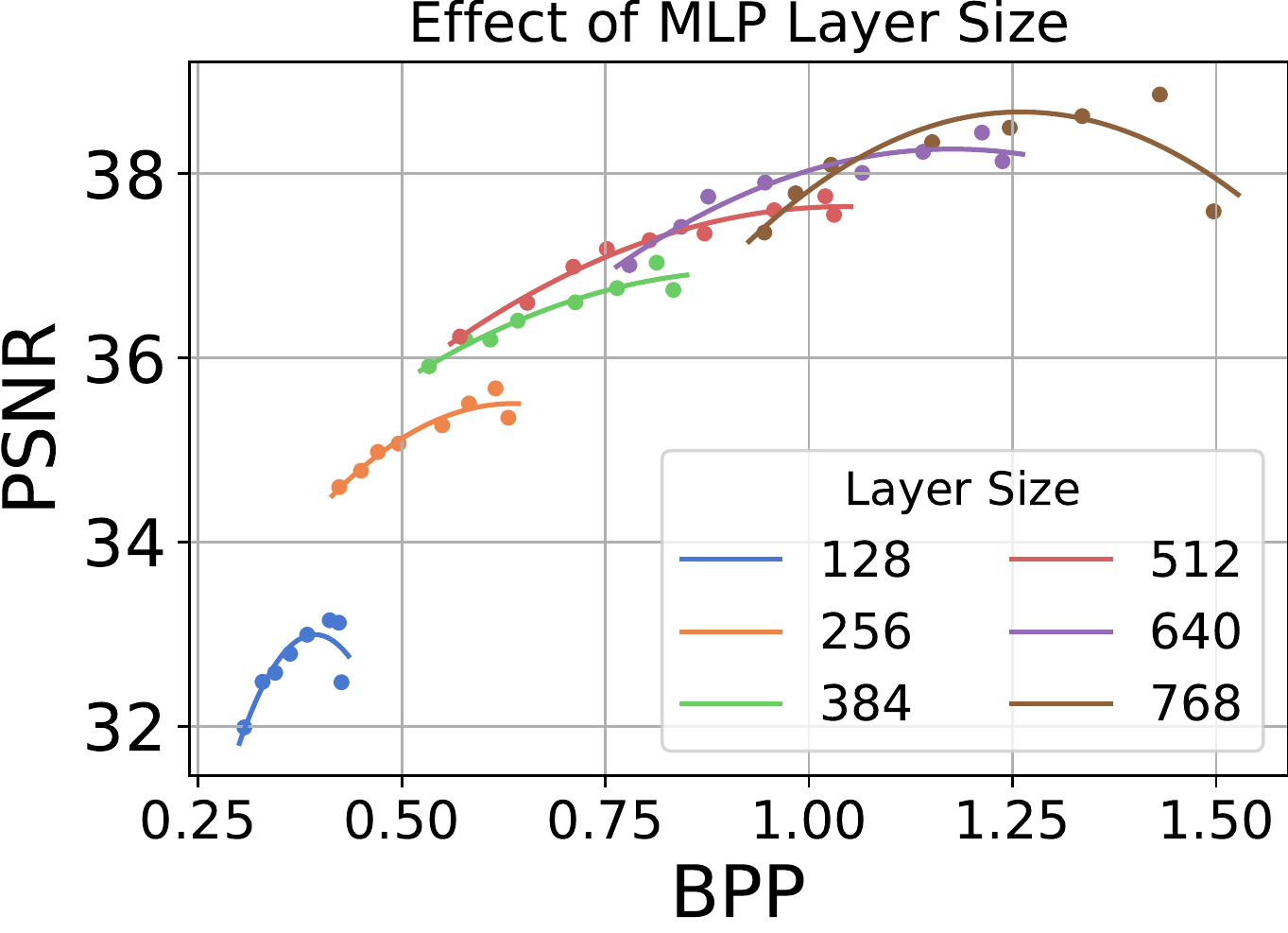}\\
    \qquad\quad(a)&\qquad\quad(b)
    \end{tabular}
    \vspace{-0.15in}
    \caption{Increasing number of layers improves the PSNR/BPP curve upto 5 layers in the lower BPP regime ($<$0.8). Increasing layer size shifts the PSNR/BPP curve upwards and to the right as representation capacity increases along with more parameters.}
    \vspace{-0.1in}
    \label{fig_supp:ablation}
\end{figure*}

\subsection{Youtube-8M}
We now provide video-wise results of 5 videos picked from the Youtube-8M dataset at 1080p resolution. Details of the videos are provided in Table~\ref{youtube8m}. Results are summarized in Table~\ref{tab_supp:ym_long_videowise}. We see that our reconstruction quality does not degrade with longer videos. This behavior is different from NeRV, which obtains a significant drop in PSNR (about $-5.2$). This is in line with the observations in Section 4.4 of the main paper, where we see that increasing number of frames results in drop of NeRV's reconstruction quality while we maintain similar levels of performance.

\renewcommand{\arraystretch}{1.1}
\begin{table}[t]
\centering
\resizebox{\linewidth}{!}{
\begin{tabular}{@{}L{\dimexpr.3\linewidth}C{\dimexpr.3\linewidth}C{\dimexpr.15\linewidth}C{\dimexpr.15\linewidth}C{\dimexpr.15\linewidth}@{}}
\toprule
\multirow{2}{*}{Video Name} & \multicolumn{2}{c}{\ours (Ours)}& \multicolumn{2}{c}{NeRV}\\
\cmidrule(l){2-3}\cmidrule(l){4-5}
&PSNR&BPP&PSNR&BPP\\
\midrule
Dota & \textbf{38.03}&0.62 &35.53 &\textbf{0.34}\\
Ride&\textbf{36.65}&1.09  &29.74  &\textbf{0.36}\\
Submarine&\textbf{38.48}&0.69 &33.64 &\textbf{0.40}\\
Water Scooter&\textbf{37.79}&0.84 &30.46 &\textbf{0.34}\\
Mortal Kombat&\textbf{36.02}&1.03 &31.36 &\textbf{0.45}\\
\midrule
Average&\textbf{37.39}&0.85 &32.14 &\textbf{0.38}\\
\bottomrule
\end{tabular}
} 
\caption{\textbf{Results on Youtube-8M videos with long duration.}
We provide video-wise results on 5 videos picked from the Youtube-8M datasets with approximately 4000 frames compared to the typical 600 from UVG. Still we maintain PSNR/BPP with no change in hyperparameters, whereas NeRV shows large degradation in performance for the same network and similar encoding times. 
}
\label{tab_supp:ym_long_videowise}
\vspace{-0.2in}
\end{table}

\section{Additional Ablations}

In addition to the ablations shown in \Fig 5, we analyze the effect of layer size and the number of layers of the MLP in our network when evaluating on the Jockey video of the UVG-HD dataset. Note that the default values of layer size is 512 and the number of layers is 5.

\subsection{Effect of Layer Number}
We increase the number of layers from 3 to 6, while keeping other parameters at their default values and varying the entropy coefficient $\lambda_I$ for each curve as in Section 5. Results are summarized in Figure~\ref{fig_supp:ablation}(a). We see that increasing the number of layers from 3 to 5 improves the tradeoff curve (shifts upwards) in the low BPP regime ($<$0.8). However, increasing it further shifts the curve upwards and to the right. This might be because the MLP network requires higher levels of non-linearity to learn a global representation for a group of 3 frames which typically contain significant motion in the case of Jockey. However, for 6 layers the network shifts the curve upwards and to the right, and we no longer obtain increase in PSNR at no cost of BPP.

\subsection{Effect of Layer Size}
We vary the layer size from 128 to 768 progressively, in steps of 128 for each of the 5 layers. Results are summarized in Figure~\ref{fig_supp:ablation}(b). We see that increasing the layer size simply shifts the curve upwards and to the right, which is expected as a higher number of parameters leads to more representation capability of the network at the cost of more parameters. While increasing the number of layers increases number of parameters as well, a similar tradeoff is not present in that case up to a certain level, suggesting that a minimum number of non-linearities/activation functions are important to achieve the optimal tradeoff.

\section{Qualitative Results}
\begin{figure*}[t]
\setlength\tabcolsep{2pt}
\centering
\begin{tabular}{ccc}
 \textbf{Ground Truth} &
 \textbf{\ours} &
 \textbf{NeRV} \\
 \includegraphics[width=0.33\textwidth]{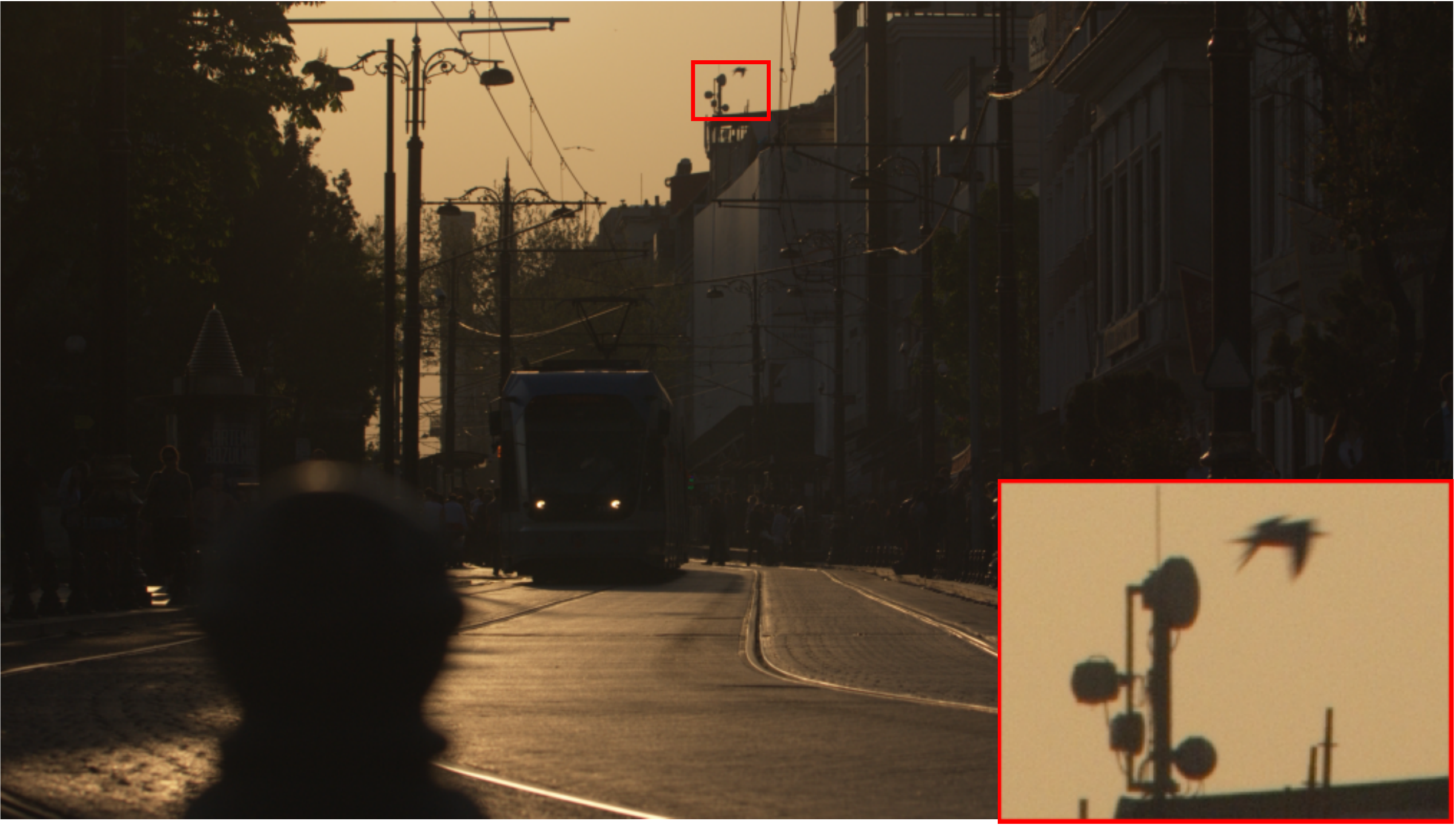} &
 \includegraphics[width=0.33\textwidth]{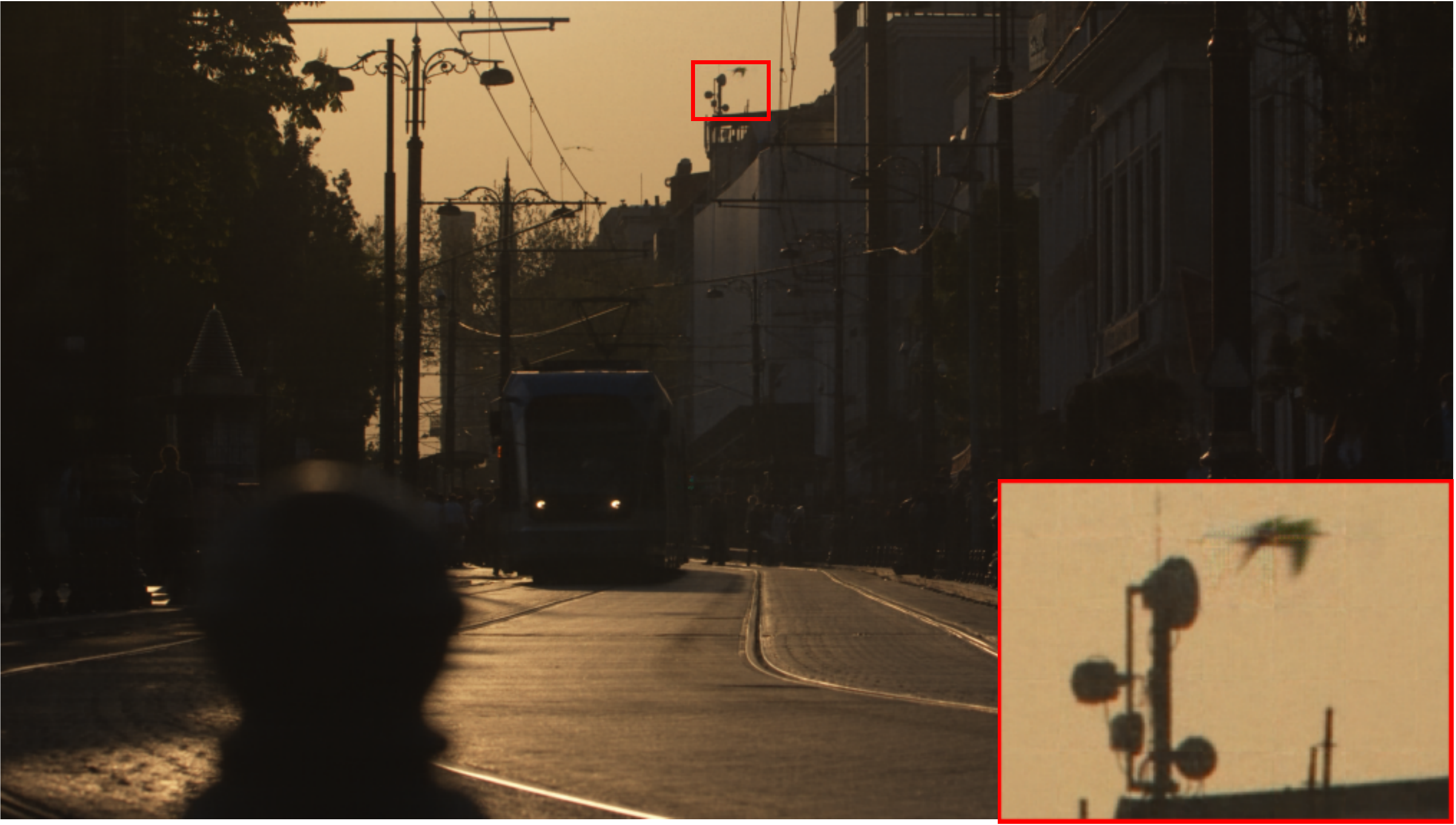} &
 \includegraphics[width=0.33\textwidth]{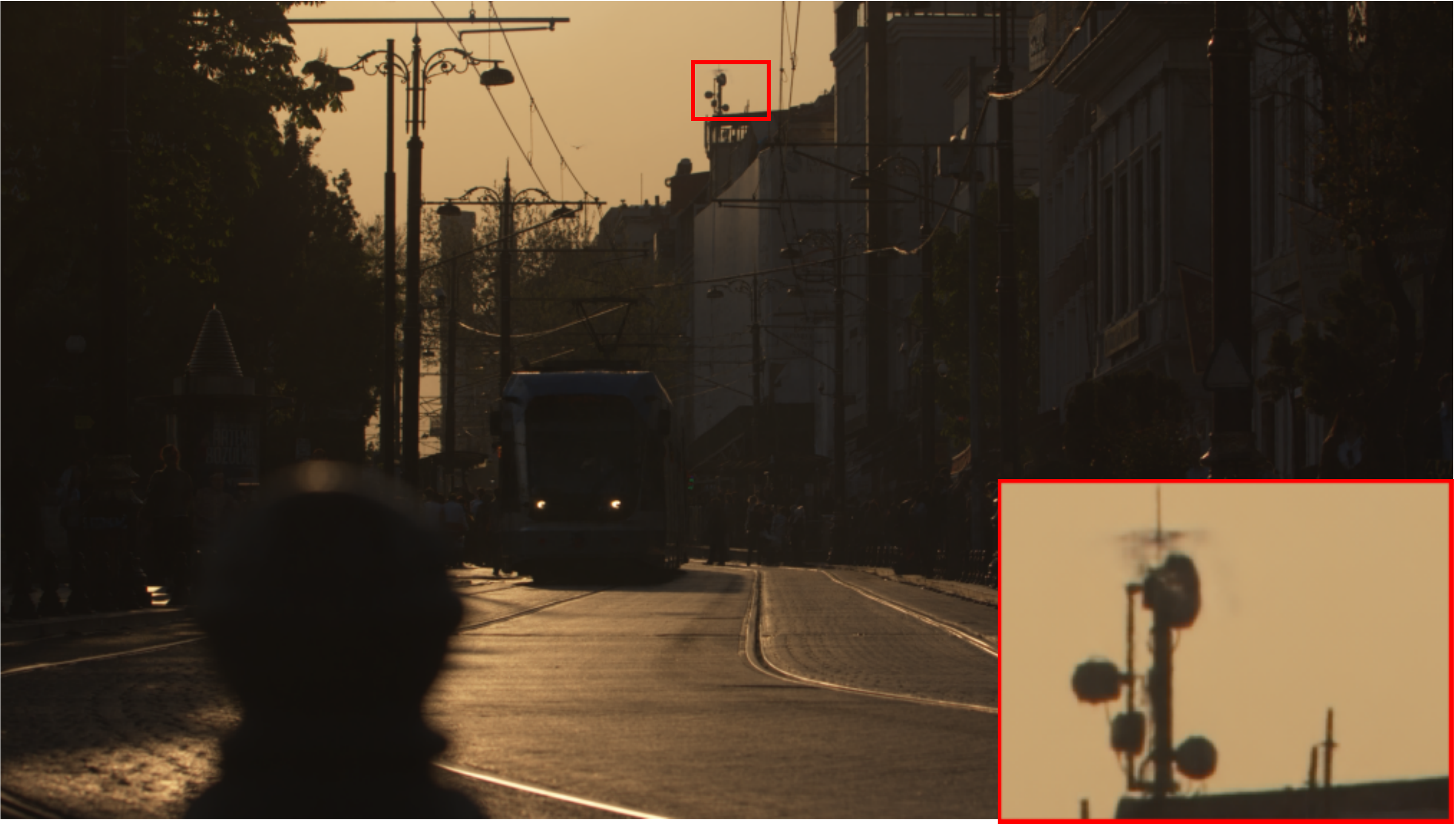} \\
 \includegraphics[width=0.33\textwidth]{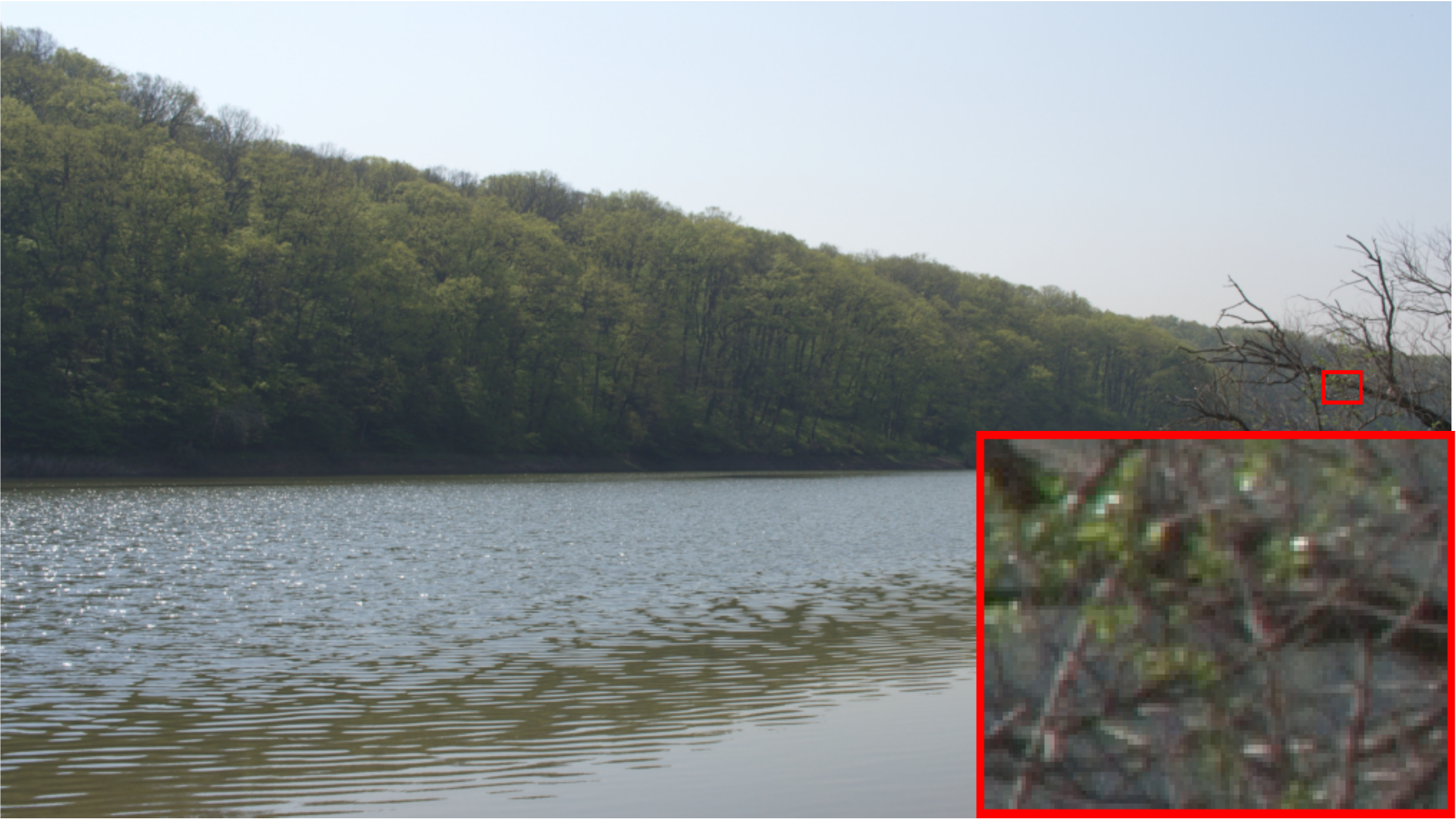} &
 \includegraphics[width=0.33\textwidth]{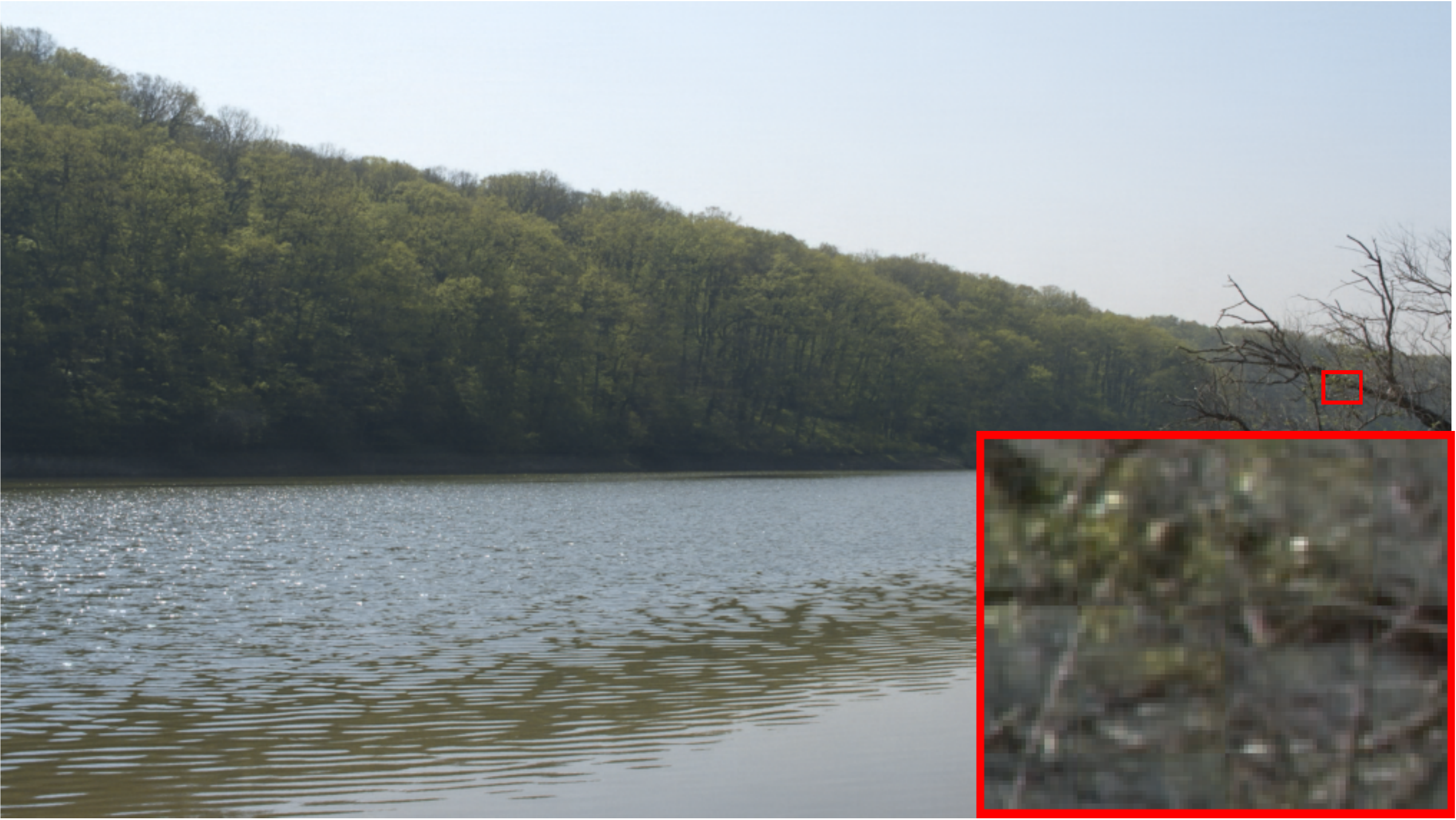} &
 \includegraphics[width=0.33\textwidth]{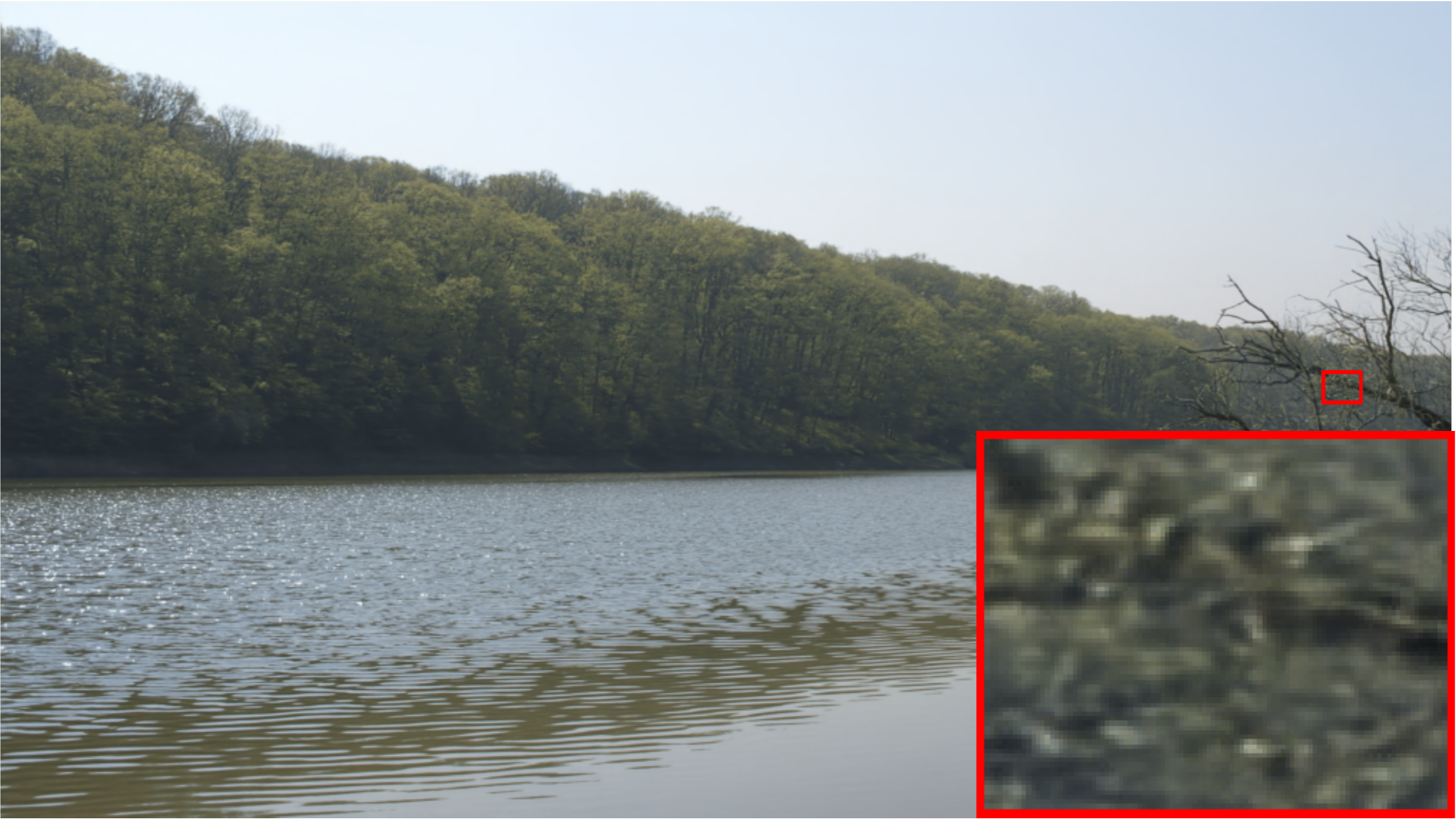} \\
 \includegraphics[width=0.33\textwidth]{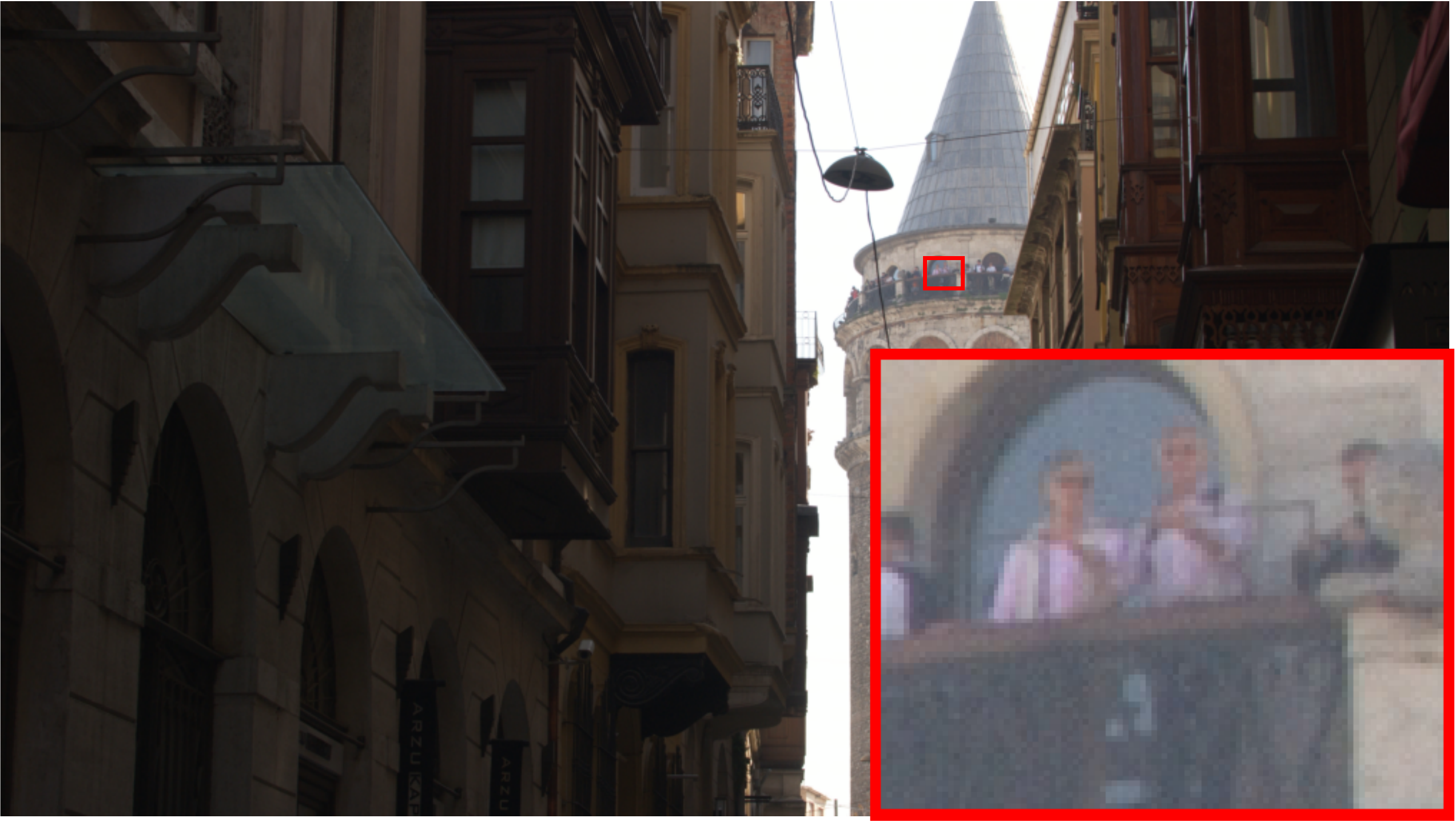} &
 \includegraphics[width=0.33\textwidth]{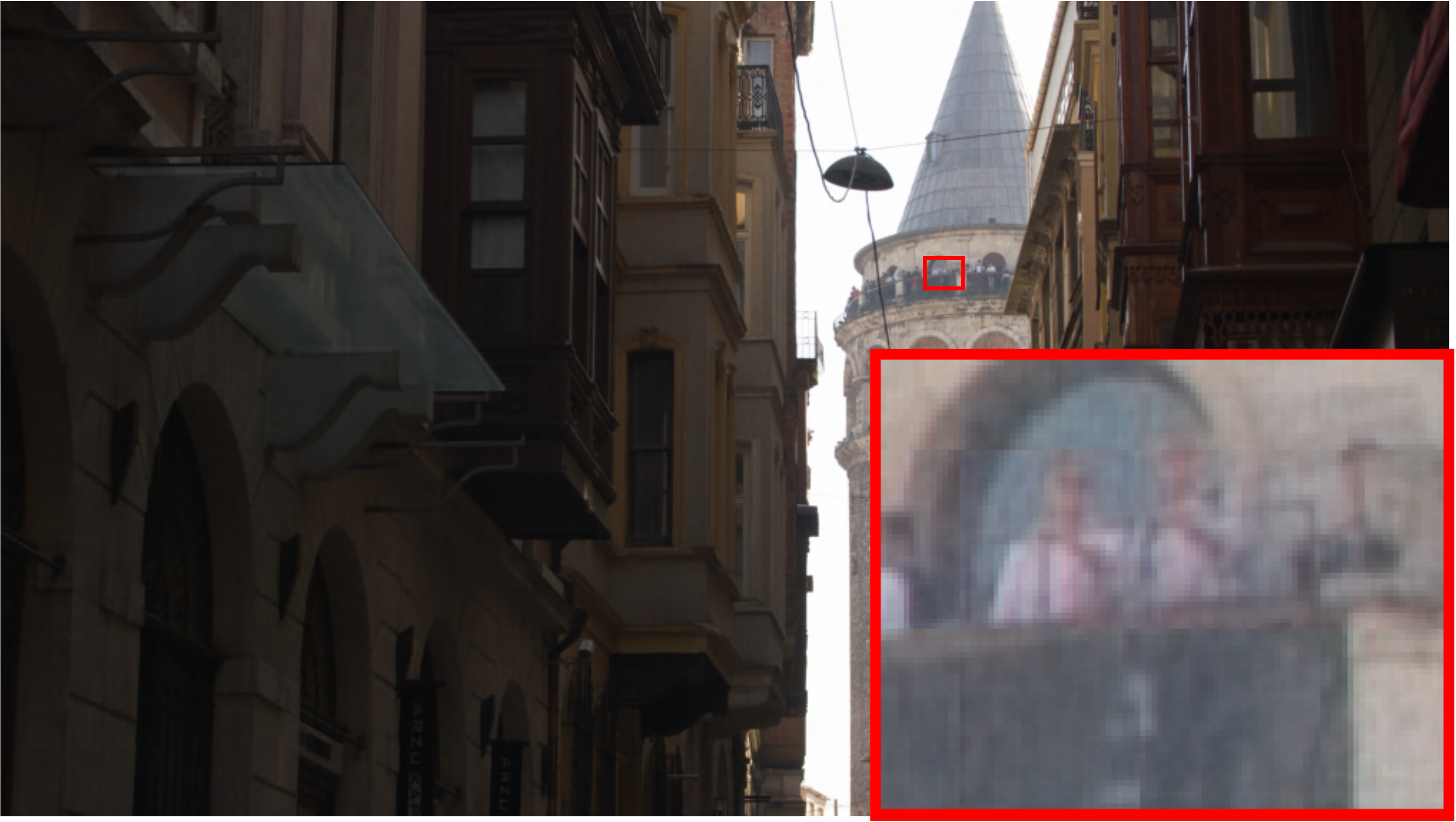} &
 \includegraphics[width=0.33\textwidth]{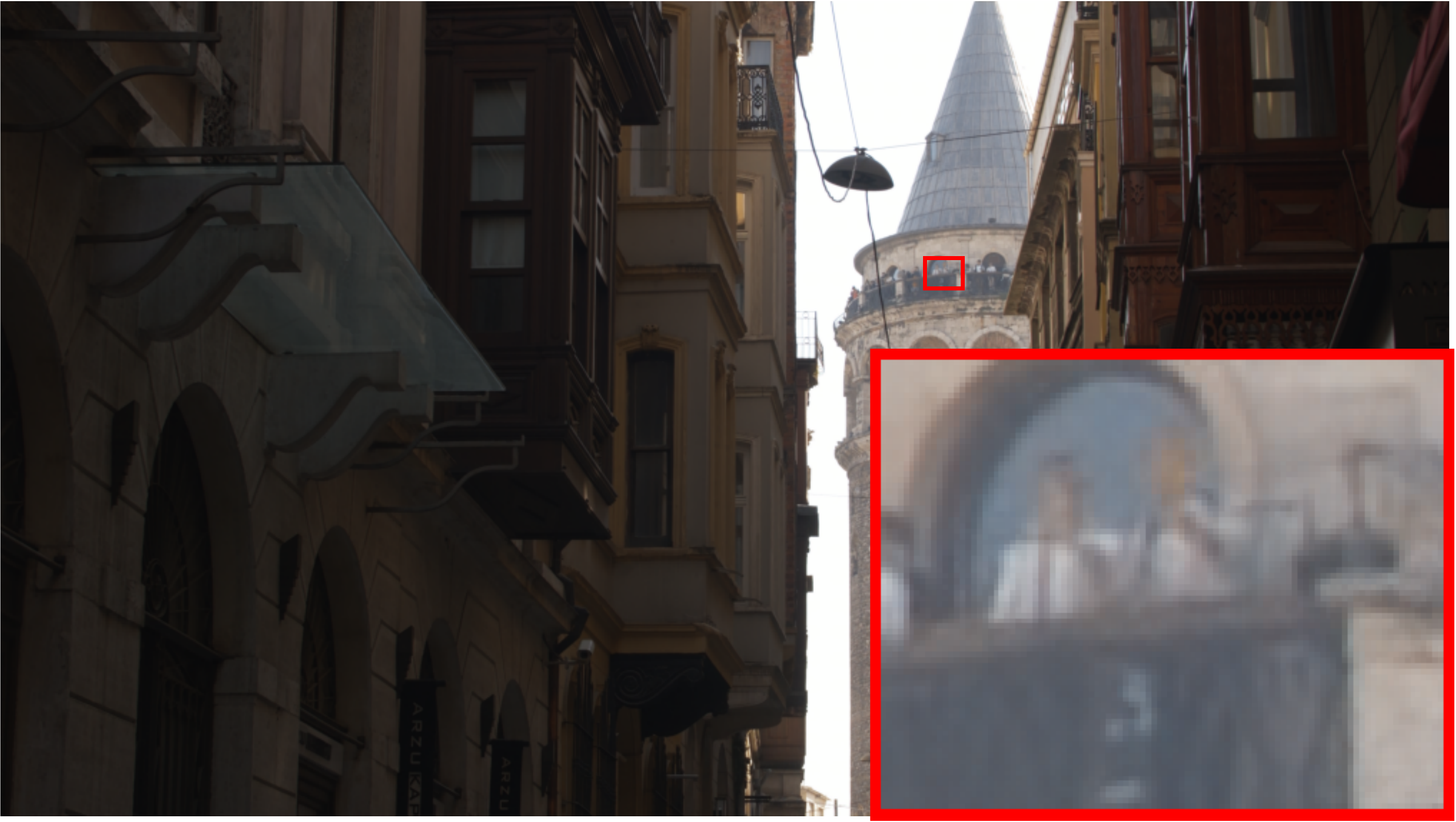}
\end{tabular}
\caption{\textbf{Qualitative results from UVG-4K Set-2:} (Left) Ground truth video frames. (Center) Reconstruction from \ours. (Right) Reconstruction from NeRV. Top to bottom: We show additional examples where \ours is able to preserve the image fidelity after reconstruction, such as the bird in Twilight, the tree in RiverBank, and the humans in CityAlley.}
\label{fig_supp:quality}
\end{figure*}

We qualitatively visualize the reconstruction results for 3 videos from UVG-4K (Set 2) in \Fig~\ref{fig_supp:quality}. We obtain higher-quality, more faithful reconstructions while preserving details at similar/lower BPP compared to NeRV; \textit{e.g.}, Twilight ($0.24\rightarrow0.13$), RiverBank ($0.24\rightarrow0.26$), CityAlley ($0.24\rightarrow0.17$). Notice the bird which is reconstructed by our approach in Twilight (top), or finer details of the branches in RiverBank (middle), or maintaining the right color information of the door and the people's shirts in CityAlley (bottom).

\clearpage


\end{document}